%% file: main.tex
\documentclass[letterpaper]{article}
\usepackage[preprint]{aaai2027}
\usepackage[hyphens]{url}
\usepackage{graphicx}
\usepackage{natbib}
\usepackage{caption}
\usepackage{algorithm}
\usepackage{algorithmic}
\usepackage{newfloat}
\usepackage{listings}
\DeclareCaptionStyle{ruled}{labelfont=normalfont,labelsep=colon,strut=off} 
\floatstyle{ruled}
\newfloat{listing}{tb}{lst}{}
\floatname{listing}{Listing}

\usepackage{booktabs}

\usepackage{microtype}
\usepackage{inconsolata}
\usepackage{multirow}
\usepackage{tabularx}
\usepackage{amsmath}
\usepackage{amssymb}
\usepackage{enumitem}
\usepackage{xspace}
\usepackage{placeins}
\usepackage{capt-of}
\usepackage{colortbl}
\usepackage{tcolorbox}
\tcbuselibrary{skins,listings}

\newcolumntype{C}{>{\centering\arraybackslash}c}
\newcolumntype{Y}{>{\centering\arraybackslash}X}

\lstdefinestyle{promptbody}{
  basicstyle=\rmfamily\footnotesize,
  numbers=none,
  breaklines=true,
  breakatwhitespace=true,
  breakindent=0pt,
  prebreak={},
  postbreak={},
  columns=fullflexible,
  keepspaces=true,
  showstringspaces=false,
  literate={-}{{-}}1,
  moredelim=**[is][\bfseries]{**}{**},
  aboveskip=0pt,
  belowskip=0pt
}

\newtcblisting{promptpanel}[2][]{
  enhanced,
  listing only,
  listing engine=listings,
  listing options={style=promptbody},
  colback=white,
  colframe=black,
  colbacktitle=white,
  boxrule=0.45pt,
  titlerule=0.45pt,
  arc=0pt,
  outer arc=0pt,
  left=1.5mm,
  right=1.5mm,
  top=1.2mm,
  bottom=1.3mm,
  title={#2},
  fonttitle=\bfseries\footnotesize,
  coltitle=black,
  before skip=2mm,
  after skip=1.5mm,
  #1
}

\newcommand{\murisbench}{\textsc{MURIS-Bench}\xspace}
\newcommand{\muris}{\textsc{MURIS}\xspace}
\newcommand{\murisinstruct}{\textsc{MURIS-Instruct}\xspace}
\newcommand{\instructsyn}{\textsc{MURIS-Instruct-Syn}\xspace}
\newcommand{\instructscan}{\textsc{MURIS-Instruct-Scan}\xspace}
\newcommand{\benchsyn}{\textsc{MURIS-Bench-Syn}\xspace}
\newcommand{\benchscan}{\textsc{MURIS-Bench-Scan}\xspace}

\newcommand{\best}[1]{\textbf{#1}}
\newcommand{\runnerup}[1]{\underline{#1}}
\newcommand{\thirdbest}[1]{\textit{#1}}

\title{MURIS: A Map-Centric Multimodal Model and Benchmark for Indoor Spatial Reasoning}
\author{
    Ming-Ming Yu\textsuperscript{\rm 1,2},
    Yu Bai\textsuperscript{\rm 2},
    Yanghong Mei\textsuperscript{\rm 3,4},
    Ziyi Bai\textsuperscript{\rm 2},\\
    Yi Chen\textsuperscript{\rm 3,4},
    Guiyu Zhao\textsuperscript{\rm 3,4},
    B\"{o}rje F. Karlsson\textsuperscript{\rm 2}\corresponding,
    Wenjun Wu\textsuperscript{\rm 1,5}\corresponding
}
\affiliations{
    \textsuperscript{\rm 1}Beihang University \quad
    \textsuperscript{\rm 2}Beijing Academy of Artificial Intelligence (BAAI)\\
    \textsuperscript{\rm 3}Institute of Automation, Chinese Academy of Sciences \quad
    \textsuperscript{\rm 4}University of Chinese Academy of Sciences\\
    \textsuperscript{\rm 5}Hangzhou International Innovation Institute, Beihang University
}

\begin{document}
\maketitle

\begin{figure*}[t]
\centering
\includegraphics[width=1.0\textwidth]{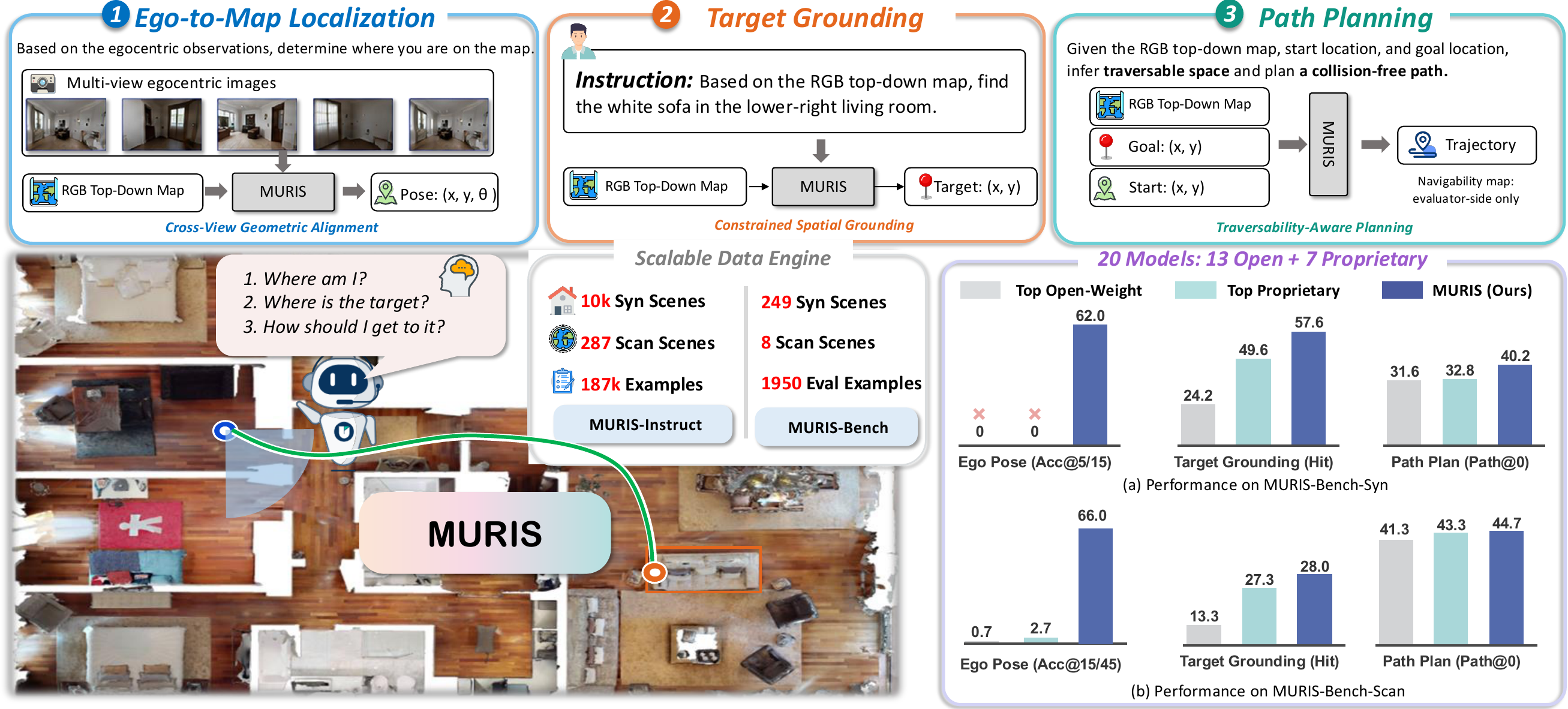}
\caption{
Overview of \murisbench and \murisinstruct. The benchmark evaluates ego-to-map localization, language-based target grounding, and map-conditioned path planning through continuous-coordinate prediction. For each benchmark split, the bottom-right panel compares \muris with the open-weight and proprietary models achieving the highest Overall Avg. across the three tasks.
}

\label{fig:task-overview}
\vspace{-5mm}
\end{figure*}

\begin{abstract}
Maps provide compact spatial memory for multi-room exploration and long-horizon navigation, yet current spatial intelligence benchmarks predominantly focus on egocentric observations or reduce maps to lossy text and discrete choices.
To advance \textbf{M}ap \textbf{U}nderstanding and \textbf{R}easoning in \textbf{I}ndoor \textbf{S}paces (\muris), we introduce \murisbench, a benchmark for direct visual-map reasoning through three continuous-coordinate tasks: ego-to-map localization (where am I), spatial language grounding (where to go), and path planning (how to get there).
To support data-driven map reasoning, we develop a scalable pipeline that converts simulator and scan metadata into geometrically verified supervision, producing \murisinstruct with 186,646 synthetic and scan-derived examples.
We evaluate 20 open-weight and proprietary MLLM configurations across synthetic and scan-derived environments, establishing a broad capability profile for direct visual-map reasoning. General-purpose MLLMs retain near-zero strict pose accuracy, while coarse semantic grounding and approximate route generation do not consistently translate into metric precision or collision-free execution. 
Controlled same-backbone experiments with \muris further show that map-specific supervision can substantially improve these capabilities without task-specific architectural modules, while mixed-source training preserves synthetic performance and strengthens scan-domain generalization. 
Data-scale and difficulty analyses show that localization improves rapidly with supervision, while instance ambiguity and obstacle-constrained geometry remain the main grounding and planning challenges.
\end{abstract}

\input{secs/intro}
\input{secs/related_works}
\input{secs/data_engine.tex}
\input{secs/method.tex}
\input{secs/experiments}
\input{secs/conclusion}

\bibliography{aaai2027}

\appendix
\setcounter{secnumdepth}{1}
\setcounter{figure}{0}
\setcounter{table}{0}
\setcounter{equation}{0}
\renewcommand{\thefigure}{S\arabic{figure}}
\renewcommand{\thetable}{S\arabic{table}}
\renewcommand{\theequation}{S\arabic{equation}}

\setcounter{topnumber}{3}
\setcounter{dbltopnumber}{3}
\renewcommand{\topfraction}{0.95}
\renewcommand{\dbltopfraction}{0.95}
\renewcommand{\textfraction}{0.04}
\renewcommand{\floatpagefraction}{0.82}
\renewcommand{\dblfloatpagefraction}{0.82}
\makeatletter
\setlength{\@dblfptop}{0pt}
\setlength{\@dblfpsep}{12pt plus 2fil}
\setlength{\@dblfpbot}{0pt plus 1fil}
\makeatother

\input{secs/appendix}

\end{document}

%% file: secs/intro.tex
\section{Introduction}

Spatial reasoning in open environments is a fundamental capability for embodied AI. Currently, research primarily centers on understanding room layouts or object relations via egocentric visual inputs. Although recent Multimodal Large Language Models (MLLMs) for spatial reasoning~\citep{zhou2026roborefer, yuan2024robopoint} have improved spatial target identification, their reliance on a restricted egocentric view confines the agent's perception to localized environmental fragments. Such a limitation creates a significant gap when tackling long-horizon, multi-room tasks, which require a macro-level grasp of global spatial structures. This local-view restriction ultimately limits reliable planning and navigation across complex, large-scale indoor environments.

To break this egocentric bottleneck, top-down maps serve as a useful structural representation. Maps provide agents with indispensable global context and spatial priors, facilitating complex behaviors such as multi-room navigation, long-term path planning, and active exploration. While specialized systems leverage maps for language-based localization~\citep{hahn2020you} or floor-plan-assisted navigation~\citep{li2025flona}, existing foundation-model pipelines often compress visual maps into lossy text or discrete candidate choices. Can MLLMs reason directly over continuous top-down map geometry for perception and planning?

To address this question, we study \textbf{M}ap \textbf{U}nderstanding and \textbf{R}easoning in \textbf{I}ndoor \textbf{S}paces (\muris) and introduce \murisbench. As shown in Figure~\ref{fig:task-overview}, the benchmark evaluates spatial reasoning on indoor maps across \benchsyn and \benchscan. Departing from traditional benchmarks that rely on lossy text translations or discrete multiple-choice questions~\citep{rodionov2025floorplanqa}, \murisbench reformulates spatial reasoning into three core tasks essential for the embodied agent loop: 1) \textit{ego-to-map localization}, answering ``where am I'' by aligning an egocentric view with a global map; 
2) \textit{language target grounding}, determining ``where to go'' by resolving multi-granularity spatial constraints to precise map coordinates to ensure unique referent identification;
and 3) \textit{waypoint path planning}, determining ``how to get there'' by jointly inferring traversable space from a top-down RGB map and generating a waypoint route. Crucially, \murisbench enforces strict coordinate and geometric verification, requiring continuous normalized-coordinate outputs rather than verbal heuristics.

To support map-grounded spatial reasoning alongside the benchmark, we develop a scalable data generation pipeline.
Using this pipeline, we construct \murisinstruct, a multimodal instruction-tuning dataset comprising \instructsyn from simulated environments and \instructscan from real-world scans.
The resulting supervision enables VLMs to learn directly from visual maps.
We instantiate \muris, a parameter-efficient map-reasoning specialist built on Qwen3-VL-4B, as a controlled reference model for evaluating benchmark learnability and cross-domain transfer.
\muris retains the standard VLM architecture and learns localization, object grounding, and waypoint planning through a unified continuous-coordinate interface, allowing performance gains to be attributed directly to \murisinstruct.

Our experiments address four questions. First, we evaluate 20 open-weight and proprietary MLLM configurations zero-shot to determine which map-centric capabilities are already present in general-purpose models. Second, controlled same-backbone studies measure how map-specific supervision improves localization, grounding, and planning, and how supervision-source composition affects cross-domain generalization. Third, difficulty-stratified and qualitative analyses identify the semantic and geometric factors shaping post-supervision performance. Fourth, data-budget experiments compare how efficiently the three capabilities improve with additional supervision. Together, these experiments establish the current capability boundary and reveal how supervision scale, source composition, semantic ambiguity, and geometric complexity shape reliable visual-map reasoning.

Our main contributions are summarized as follows:
(1) We formulate three continuous-coordinate tasks and introduce \murisbench, a unified benchmark for evaluating localization, language target grounding, and path planning directly on top-down indoor maps.
(2) We develop a scalable synthetic and scan-derived data engine with deterministic geometric verification, and construct the \murisinstruct training dataset.
(3) We conduct a broad evaluation of 20 open-weight and proprietary MLLM configurations across synthetic and scan-derived maps, revealing that coarse semantic and routing priors do not yield reliable metric localization, precise target grounding, or collision-free planning.
(4) Controlled same-backbone, source-composition, difficulty-stratified, and data-budget studies reveal distinct map-reasoning acquisition regimes: mixed-source supervision yields a unified model with stronger cross-domain performance, localization scales strongly with supervision, and diagnostic analyses identify instance disambiguation and obstacle-constrained geometry as the main grounding and planning challenges.

%% file: secs/related_works.tex
\section{Related Work}

\paragraph{General spatial reasoning in large models.}
Recent work evaluates and improves spatial reasoning in Multimodal Large Language Models (MLLMs) \citep{xing2025can, jia2025omnispatial, ma20253dsrbench, stogiannidis2025mind, wang2025spatial457, zheng2025multimodal}. 
Benchmarks like Cambrian-S~\citep{yang2025cambrian} and OST-Bench \citep{lin2026ost} evaluate macro spatio-temporal relations, while works such as RoboRefer \citep{zhou2026roborefer} and RoboPoint \citep{yuan2024robopoint} target fine-grained, point-level spatial grounding. Parallel to these evaluations, recent methods attempt to augment MLLMs' spatial acuity via metric supervision, depth representations \citep{bhat2023zoedepth}, or explicit reasoning traces \citep{chen2024spatialvlm, cheng2024spatialrgpt, cai2025spatialbot, liao2024reasoning, wu2024mind} across foundational model families \citep{achiam2023gpt, comanici2025gemini, anthropic2026sonnet46, liu2024improved, bai2025qwen2,liu2023visual, liu2024llavanext}. 
However, both benchmarks and methods largely focus on egocentric images or video, leaving top-down visual-map reasoning underexplored.

\paragraph{Map perception and understanding.}
To reason beyond immediate ego-views, maps provide structured representations of scene layout and semantics. Prior work has studied floorplan reconstruction and room-region parsing from RGB-D scans, point clouds, or projected density maps, including FloorNet~\citep{liu2018floornet}, Floor-SP~\citep{chen2019floor}, Scan2Plan~\citep{phalak2020scan2plan}, HEAT~\citep{chen2022heat}, and RoomFormer~\citep{yue2023connecting}. 
Another line builds semantic or open-vocabulary maps from egocentric RGB-D observations or video streams for language-conditioned querying and navigation, such as NLMap~\citep{chen2023open}, VLMaps~\citep{huang2023visual}, ConceptGraphs~\citep{gu2024conceptgraphs}, and OpenMap~\citep{li2025openmap}. 
These works advance layout reconstruction, semantic map construction, and language-conditioned retrieval, but they do not directly evaluate whether general-purpose MLLMs can reason over top-down visual maps and produce continuous metric outputs for localization, grounding, and path planning.

\paragraph{Map-grounded embodied AI.}
Beyond map-free embodied policies relying solely on egocentric observations~\citep{fu2024mobile, zhang2024navid, yu2025cnav, meri2026}, explicit maps are widely used in embodied AI to support long-horizon navigation, object search, and interaction with physical environments.
Earlier work has connected embodied localization with map grounding through dialog. WAY~\citep{hahn2020you} localizes an agent on a map from embodied dialog, showing that language interactions can provide useful spatial cues for recovering the agent's position. This setting highlights the importance of map-level localization, but its grounding signal is primarily linguistic rather than a direct alignment between egocentric visual observations and a top-down visual map. 
More recent zero-shot embodied frameworks explicitly construct environmental maps and leverage LLMs or VLMs for high-level semantic reasoning and decision-making~\citep{gadre2023cows, kuang2024openfmnav, yu2023l3mvn, wu2024voronav, zhou2023esc, yu2026ranger}. 
However, these systems often expose maps to language models as textual or discretized summaries and reduce navigation decisions to selecting among heuristic waypoint candidates, rather than predicting continuous metric coordinates directly from visual map structure.
This input-side textualization and output-side discretization can obscure fine geometric details needed for physical grounding.
Evaluating whether MLLMs can reason directly over top-down visual maps and produce metric outputs is therefore a necessary step toward map-grounded embodied intelligence.

%% file: secs/data_engine.tex
\section{\murisbench and Data Engine}

\subsection{Design Principles}

\textbf{Input-side native map reasoning.} Unlike methods that compress spatial layouts into text, \murisbench requires models to reason directly over top-down RGB maps $M \in \mathbb{R}^{H \times W \times 3}$. This preserves environmental context and continuous spatial geometry lost through text translation.

\textbf{Output-side coordinate prediction.} Instead of formulating navigation decisions as classification over discrete heuristic point candidates, we unify localization, grounding, and planning as predicting continuous 2D coordinates within a normalized space $\Omega=[0,1000]^2$. This formulation provides a universal geometric interface between vision-language reasoning and downstream embodied controllers.

\subsection{Core Tasks and Evaluation Metrics}
\label{sec:tasks-metrics}

\begin{figure*}[t]
\centering
\includegraphics[width=1.0\textwidth]{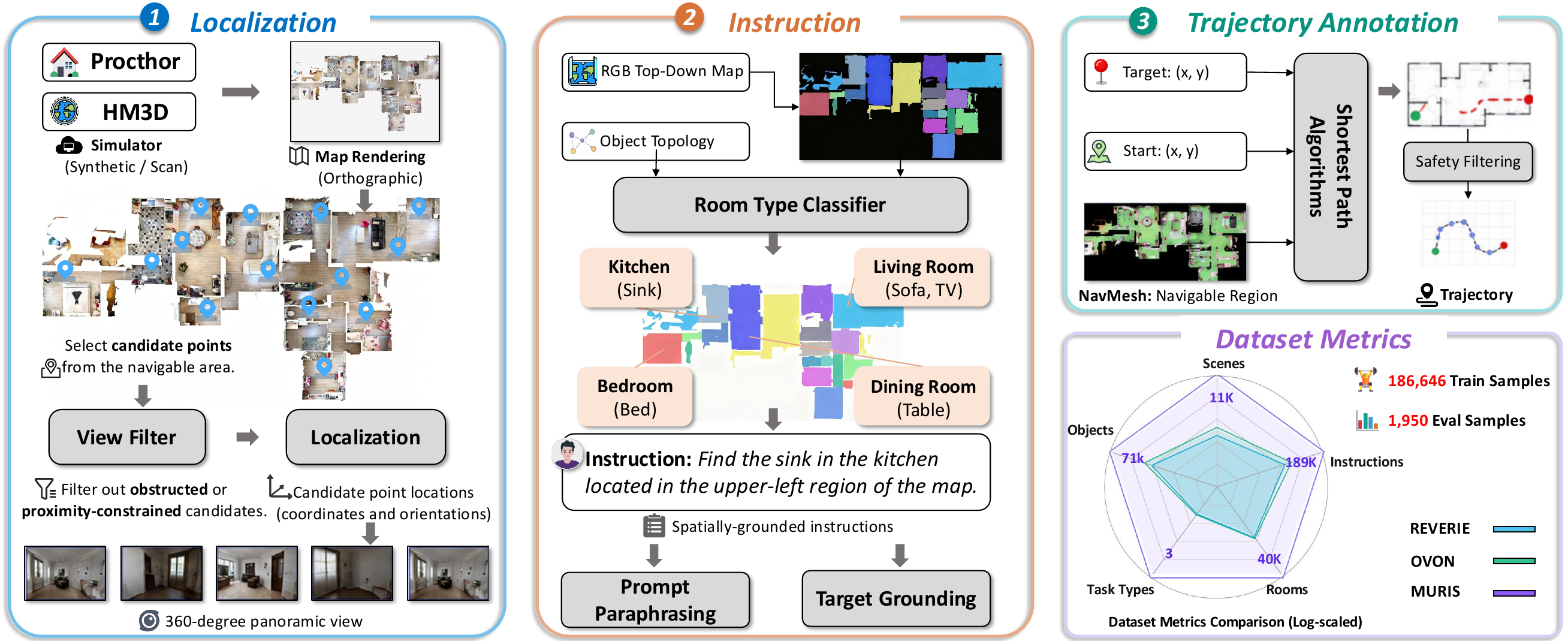}
\caption{Data construction pipeline for \murisinstruct and \murisbench. Simulator and scan metadata are converted into cross-view pose labels, uniquely grounded spatial instructions, and geometrically valid waypoint paths. The bottom-right panel compares overall dataset scale against representative indoor navigation benchmarks (REVERIE and OVON).}
\label{fig:data-pipeline}
\vspace{-4mm}
\end{figure*}

\begin{table*}[t]
  \centering
  \scriptsize
  \setlength{\tabcolsep}{3pt}
  \renewcommand{\arraystretch}{0.88}
  \resizebox{\textwidth}{!}{%
    \begin{tabular}{CCCCCCCCCCC}
      \toprule
      \multirow{2}{*}{Model} &
      \multicolumn{3}{c}{Task~1 Pose} &
      \multicolumn{2}{c}{Task~2 Grounding} &
      \multicolumn{4}{c}{Task~3 Path} &
      \multicolumn{1}{c}{Overall} \\
      \cmidrule(lr){2-4}\cmidrule(lr){5-6}\cmidrule(lr){7-10}
      & Mean L2$\downarrow$ & Acc@5/15$\uparrow$ & Acc@10/30$\uparrow$ &
      Hit$\uparrow$ & Acc@10$\uparrow$ & Path@0$\uparrow$ & SPL$\uparrow$ & Coll. Ratio$\downarrow$ & Med. DTW$\downarrow$ & Avg.$\uparrow$ \\
      \midrule
      \multicolumn{11}{l}{\textbf{\textit{Open-weight models}} \textit{(MoE: total/active)}} \\
      InternVL3.5-8B~\citep{wang2025internvl3} & 108.18 & \runnerup{0.2} & 0.2 & 9.0 & 13.2 & 23.0 & 21.03 & 50.55 & 297.06 & 10.7 \\
      InternVL3.5-38B~\citep{wang2025internvl3} & 98.63 & 0.0 & 0.0 & \thirdbest{24.2} & 28.2 & 31.6 & 30.85 & 20.36 & 112.66 & \thirdbest{18.6} \\
      Molmo-7B-D~\citep{deitke2025molmo} & 100.52 & 0.0 & 0.2 & 6.0 & 9.8 & 3.6 & 3.31 & 37.09 & 818.59 & 3.2 \\
      Molmo-72B-0924~\citep{deitke2025molmo} & 103.40 & 0.0 & 0.0 & 14.6 & 25.2 & 25.8 & 23.06 & 28.12 & 160.11 & 13.5 \\
      RoboBrain2.5-8B~\citep{tan2026robobrain} & 114.92 & \runnerup{0.2} & 0.4 & 11.2 & 15.0 & 17.4 & 15.76 & 22.54 & 208.71 & 9.6 \\
      MolmoPoint-8B~\citep{clark2026molmopoint} & 133.49 & 0.0 & 0.0 & 2.8 & 3.8 & 30.8 & 27.70 & 27.63 & 197.92 & 11.2 \\
      LLaVA-OneVision-2-8B~\citep{an2026llava} & 147.13 & 0.0 & 0.2 & 3.2 & 3.6 & 24.8 & 22.97 & 40.54 & 272.23 & 9.3 \\
      Qwen3-VL-4B~\citep{bai2025qwen3} & 97.09 & 0.0 & 0.0 & 12.8 & 19.0 & 29.2 & 26.53 & 24.19 & 114.48 & 14.0 \\
      Qwen3-VL-8B~\citep{bai2025qwen3} & 102.29 & \runnerup{0.2} & 0.2 & 5.6 & 10.0 & 31.8 & 28.47 & 23.79 & 112.25 & 12.5 \\
      Qwen3-VL-32B~\citep{bai2025qwen3} & 91.57 & 0.0 & 0.0 & 13.8 & 21.8 & 30.0 & 27.41 & 21.75 & 122.15 & 14.6 \\
      GLM-4.6V (106B/12B)~\citep{hong2025glm} & 104.13 & 0.0 & 0.0 & 7.4 & 13.8 & \runnerup{37.4} & \runnerup{36.79} & 19.33 & \thirdbest{91.51} & 14.9 \\
      Kimi-K2.6 (1.04T/32B)~\citep{moonshot2026kimi26} & 88.54 & 0.0 & \thirdbest{0.6} & 3.6 & 8.0 & 30.8 & 27.46 & 23.03 & 135.38 & 11.5 \\
      MiMo-V2.5 (310B/15B)~\citep{xiaomi2026mimov25} & 95.49 & 0.0 & 0.2 & 4.8 & 8.0 & \thirdbest{34.4} & 31.86 & 21.24 & 104.72 & 13.1 \\
      \midrule
      \multicolumn{11}{l}{\textbf{\textit{Proprietary models}}} \\
      Qwen3-VL-Plus~\citep{bai2025qwen3} & 109.38 & 0.0 & 0.4 & 18.8 & 27.0 & 32.0 & 29.80 & 20.86 & 111.08 & 16.9 \\
      Gemini-2.5-Flash~\citep{comanici2025gemini} & 73.51 & 0.0 & 0.0 & 4.0 & 8.0 & 26.8 & 24.13 & 20.40 & 154.04 & 10.3 \\
      Gemini-2.5-Pro~\citep{comanici2025gemini} & 69.95 & 0.0 & 0.2 & 20.4 & \thirdbest{31.4} & 32.4 & 30.71 & 18.63 & \runnerup{90.42} & 17.6 \\
      GPT-5~\citep{singh2025openai} & \thirdbest{69.91} & 0.0 & \thirdbest{0.6} & 3.6 & 6.0 & 32.6 & 30.81 & 18.54 & 154.87 & 12.1 \\
      GPT-5.4~\citep{openai2026gpt54} & 82.03 & \runnerup{0.2} & \runnerup{1.0} & 7.6 & 12.2 & 34.0 & \thirdbest{32.00} & \thirdbest{18.38} & 104.60 & 13.9 \\
      Claude-Sonnet-4.6~\citep{anthropic2026sonnet46} & 91.62 & 0.0 & 0.4 & 12.8 & 18.8 & 31.2 & 29.72 & 21.62 & 121.62 & 14.7 \\
      Doubao-Seed-2.0-Pro~\citep{seed2026seed2} & \runnerup{60.13} & 0.0 & \thirdbest{0.6} & \runnerup{49.6} & \runnerup{52.8} & 32.8 & 29.06 & \runnerup{18.01} & 102.19 & \runnerup{27.5} \\
      \midrule
      \muris & \best{5.22} & \best{62.0} & \best{92.2} & \best{57.6} & \best{65.8} & \best{40.2} & \best{39.17} & \best{14.99} & \best{69.10} & \best{53.3} \\
      \bottomrule
    \end{tabular}%
  }
  \vspace{-1mm}
  \caption[MURIS-Bench-Syn averaged-split results]{\benchsyn aggregate results. Accuracy metrics and collision ratios are percentages; Overall Avg. averages Acc@5/15, Hit, and Path@0. Bold, underlined, and italic values mark the top three non-zero results.}
  \label{tab:syn-average}
  \vspace{-4mm}
\end{table*}

\paragraph{Task 1: Ego-to-Map Localization}
This task tests whether local first-person observations can be registered to a global map frame. Formally, given a top-down map $M$ and five ordered egocentric RGB views $E = (e_0, \dots, e_4)$ from the same camera location, the model maps $f_{\mathrm{loc}}(M, E) \rightarrow (\hat{p}, \hat{\theta})$, where $p \in \Omega$ is the shared normalized position and $\theta \in [0,360)$ is the heading of $e_4$.

\textbf{Evaluation Metrics.} Across all tasks, $d_2(\cdot,\cdot)$ denotes pixel (px) distance after mapping from $\Omega$ to the native $H\times W$ raster. With circular heading error $d_{\theta}(\hat{\theta}, \theta) = \min\!\left(|\hat{\theta}-\theta|,\,360-|\hat{\theta}-\theta|\right)$ in $[0,360)$, joint pose accuracy $\mathrm{Pose}@(r, \tau)$ ($\mathrm{Acc}@r/\tau$ in tables, for $r$~px and $\tau^\circ$) is:
\begin{equation}
\mathrm{Pose}@(r, \tau) = \frac{1}{N} \sum_i \mathbf{1}_{\{ d_2(\hat{p}_i, p_i) \le r,\; d_{\theta}(\hat{\theta}_i, \theta_i) \le \tau \}}.
\end{equation}

\paragraph{Task 2: Language Target Grounding}
This task evaluates complex spatial instruction following. Given a map $M$ and a natural-language query $q$ encoding multi-step spatial constraints (e.g., room context, topological relations, and distractor elimination), the model predicts the target object center: $f_{\mathrm{grd}}(M, q) \rightarrow \hat{p}^{\star} \in \Omega$.

\textbf{Evaluation Metrics.} Spatial precision uses thresholded accuracy $\mathrm{Acc}@r = \frac{1}{N}\sum_i \mathbf{1} (d_2(\hat{p}^{\star}_i, p^{\star}_i) \leq r)$ ($r=10$~px), while object-level hit accuracy (\(\mathrm{Hit}\)) checks whether $\hat{p}^{\star}_i$ falls within the visible region mask or bounding box \(\mathcal{R}_i\) of the target instance.

\paragraph{Task 3: Map-Conditioned Path Planning}
This task evaluates joint visual traversability reasoning and waypoint generation from a top-down RGB map. Given only the RGB map $M$, a start coordinate $s \in \Omega$, and a goal coordinate $g \in \Omega$, the model must infer traversable space from visual appearance, including walls and furniture such as beds and sofas, and generate a waypoint sequence $\hat{P}=(\hat{p}_1, \dots, \hat{p}_m)$ connecting them: $f_{\mathrm{plan}}(M, s, g) \rightarrow \hat{P}$. The paired navigability map derived from simulator or scan geometry is reserved for trajectory construction and collision-based evaluation.

\textbf{Evaluation Metrics.} We rasterize the complete predicted polyline in the native map raster and denote its traversed pixel set by $\mathcal{R}(\hat P)$ and its obstacle or out-of-bounds subset by $\mathcal{C}(\hat P)$. A predicted path is successful under Path@0 only when it is parsed with at least two waypoints, its first and last waypoints satisfy $d_2(\hat{p}_1,s) \le \epsilon_s$ and $d_2(\hat{p}_m,g) \le \epsilon_g$ ($\epsilon_s=\epsilon_g=5$~px, ${\approx}0.20$~m), and $\mathcal{C}(\hat P)$ is empty:
\begin{equation}
\begin{aligned}
\mathrm{Path}@0 = \frac{1}{N}\sum_i \mathbf{1}\big[&\mathrm{valid}(\hat{P}_i) \land d_2(\hat{p}_{i,1},s_i)\le\epsilon_s \\
&\land d_2(\hat{p}_{i,m_i},g_i)\le\epsilon_g \land |\mathcal{C}(\hat P_i)|=0\big].
\end{aligned}
\end{equation}
Unparseable outputs or sequences with fewer than two waypoints are counted as failures. We further report collision-free success weighted by path length,
\begin{equation}
\mathrm{SPL}=\frac{1}{N}\sum_i S_i\frac{L_i}{\max(L_i,\hat{L}_i)},
\end{equation}
where $S_i$ is the Path@0 success indicator and $L_i$ and $\hat{L}_i$ are the reference and predicted path lengths in the native map raster. Thus SPL jointly requires endpoint correctness, zero collision, and an efficient route. To quantify collision severity independently of waypoint partitioning, we report the pooled collision ratio $\sum_i|\mathcal{C}(\hat P_i)|/\sum_i|\mathcal{R}(\hat P_i)|$ over parsed paths. We additionally report dynamic time warping (\(\mathrm{DTW}\)) as a secondary measure of geometric alignment with the canonical reference route. Additional rasterization details and results under relaxed collision thresholds are provided in the appendix.

\subsection{Scalable Data Engine}

To train and rigorously evaluate map understanding, models require large amounts of spatially grounded data. 
However, human annotation of pixel-perfect coordinates, complex spatial relations, and collision-free paths on top-down maps is expensive and difficult to scale. 
To address this, we introduce an automated pipeline that leverages simulator metadata and existing 3D annotations to generate precise 2D coordinate annotations, eliminating additional manual annotation.
\paragraph{Automated Annotation Pipeline.}

For both synthetic (ProcTHOR~\citep{deitke2022}) and scanned (HM3D~\citep{yadav2023habitat}) environments, we extract top-down orthographic RGB maps directly from the 3D meshes with a fixed pixel scale.

For \textit{Localization} (Task 1), we sample candidate poses from the navigable region with farthest-point sampling to encourage spatial coverage. We render panoramic RGB views at each pose and filter low-quality observations using depth statistics and the ratio of black pixels in the RGB image, removing viewpoints that are heavily occluded or affected by scan artifacts. The retained 3D camera poses are then projected onto the normalized 2D map space $\Omega$.

For \textit{Language Grounding} (Task 2), we build a spatial scene graph from object metadata, including 3D positions, categories, room or region membership, and inter-object spatial relations. Following~\citet{ramakrishnan2021hm3d}, we resolve the spatial sparsity of original HM3D room annotations by inferring region types through class-weighted voting over the enclosed object instances. We then generate uniquely grounded instructions using room anchors, directional cues, object relations, and same-category distractors.

 For \textit{Path Planning} (Task 3), we rasterize simulator navigability into a 2D traversability map that incorporates the simulator's embodied-agent footprint and clearance constraints. To avoid trivial scenarios, we constrain the sampling of start and goal points by enforcing a minimum path length and requiring cross-region placement in multi-room environments. We then run a shortest-path search on this map to obtain safe trajectories, followed by safety filtering and orthographic projection into 2D waypoint sequences. As an additional quality-control step, we conduct human spot checks on randomly sampled generated examples to verify annotation correctness and pipeline quality.

\paragraph{\murisinstruct and \murisbench.}
As shown in Figure~\ref{fig:data-pipeline}, the pipeline produces \murisinstruct for training and \murisbench for evaluation. 

\textbf{\murisinstruct} contains 186,646 training examples: \instructsyn contributes 157,378 examples from 9,960 ProcTHOR houses, and \instructscan contributes 29,268 examples from 287 HM3D scenes, spanning 10,247 training environments in total.

\textbf{\murisbench} contains 1,950 evaluation examples: \benchsyn contributes 1,500 examples from 249 unique ProcTHOR houses, and \benchscan contributes 450 examples from 8 HM3D scenes. For every task, the training and evaluation splits are scene-disjoint.
More detailed data distributions and visualization examples are provided in the appendix (\textit{Data Details}).

%% file: secs/method.tex
\section{\muris as a Data-Trained Map Specialist}

We instantiate \muris by adapting Qwen3-VL-4B with LoRA~\citep{bai2025qwen3,hu2022lora} under a controlled backbone, enabling a direct evaluation of the training utility of \murisinstruct. The primary results use a unified model trained on the union of synthetic and scan-derived supervision, while the source-composition study compares synthetic-only, scan-only, and mixed-source variants. Because the architecture is held fixed across all variants, the observed gains isolate the effect of map-specific supervision. We formulate localization, grounding, and planning as autoregressive coordinate prediction in a shared normalized output space. Task-aligned prompts provide the required visual inputs; for example, Task 1 jointly encodes a global map and five egocentric views. \muris generates coordinate text tokens within \(\Omega\) using next-token cross-entropy. Full training and optimization details are provided in the appendix.

%% file: secs/experiments.tex
\section{Experiments}

We evaluate map-centric spatial reasoning across synthetic and scan-derived environments around four questions:
(1) What map-centric localization, grounding, and planning capabilities do general-purpose MLLMs exhibit under zero-shot evaluation?
(2) To what extent can map-specific supervision improve these capabilities, and how does supervision-source composition affect cross-domain generalization under a controlled same-backbone setting?
(3) Which semantic and geometric factors shape performance after map-specific supervision?
(4) How do the three capabilities respond to increasing amounts of task-specific supervision?

\begin{table*}[t]
  \centering
  \scriptsize
  \setlength{\tabcolsep}{2.5pt}
  \renewcommand{\arraystretch}{0.88}
  \resizebox{\textwidth}{!}{%
    \begin{tabular}{CCCCCCCCCCCC}
      \toprule
      \multirow{2}{*}{Model} &
      \multicolumn{4}{c}{Task~1 Pose} &
      \multicolumn{2}{c}{Task~2 Grounding} &
      \multicolumn{4}{c}{Task~3 Path} &
      \multicolumn{1}{c}{Overall} \\
      \cmidrule(lr){2-5}\cmidrule(lr){6-7}\cmidrule(lr){8-11}
      & Mean L2$\downarrow$ & Acc@15/45$\uparrow$ & Acc@20/60$\uparrow$ & Acc@25/90$\uparrow$ &
      Hit$\uparrow$ & Acc@10$\uparrow$ & Path@0$\uparrow$ & SPL$\uparrow$ & Coll. Ratio$\downarrow$ & Med. DTW$\downarrow$ & Avg.$\uparrow$ \\
      \midrule
      \multicolumn{12}{l}{\textbf{\textit{Open-weight models}} \textit{(MoE: total/active)}} \\
      InternVL3.5-8B~\citep{wang2025internvl3} & 218.23 & 0.0 & 0.0 & 0.0 & 6.0 & 2.0 & 20.0 & 19.28 & 60.52 & 431.14 & 8.7 \\
      InternVL3.5-38B~\citep{wang2025internvl3} & 184.15 & 0.0 & 0.0 & 0.0 & 13.3 & 10.0 & 32.0 & 31.78 & 22.53 & 266.66 & 15.1 \\
      Molmo-7B-D~\citep{deitke2025molmo} & 190.94 & 0.0 & 0.0 & 0.0 & 4.7 & 3.3 & 0.7 & 0.67 & 49.55 & 2115.17 & 1.8 \\
      Molmo-72B-0924~\citep{deitke2025molmo} & 192.84 & 0.0 & 0.0 & 0.0 & 10.0 & 6.7 & 20.7 & 19.45 & 44.92 & 414.94 & 10.2 \\
      RoboBrain2.5-8B~\citep{tan2026robobrain} & 224.54 & 0.0 & 0.0 & 0.7 & 11.3 & 10.0 & 20.7 & 19.89 & 33.31 & 289.56 & 10.7 \\
      MolmoPoint-8B~\citep{clark2026molmopoint} & 242.08 & 0.7 & 0.7 & 0.7 & 0.7 & 0.7 & 22.7 & 21.70 & 36.22 & 378.67 & 8.0 \\
      LLaVA-OneVision-2-8B~\citep{an2026llava} & 172.43 & 0.0 & 0.7 & 2.0 & 6.7 & 0.0 & 16.0 & 15.17 & 49.74 & 402.58 & 7.6 \\
      Qwen3-VL-4B~\citep{bai2025qwen3} & 185.91 & 0.0 & 0.0 & 1.3 & 13.3 & 8.0 & 24.7 & 23.44 & 47.12 & 422.72 & 12.7 \\
      Qwen3-VL-8B~\citep{bai2025qwen3} & 169.34 & 0.0 & 0.0 & 1.3 & 14.0 & 6.7 & 30.0 & 28.45 & 20.90 & 359.80 & 14.7 \\
      Qwen3-VL-32B~\citep{bai2025qwen3} & 146.67 & 0.7 & 2.0 & 4.7 & 14.0 & 10.7 & 26.0 & 23.95 & 28.86 & 334.77 & 13.6 \\
      GLM-4.6V (106B/12B)~\citep{hong2025glm} & 226.20 & 0.7 & 0.7 & 2.0 & 13.3 & 10.7 & \thirdbest{41.3} & \runnerup{40.61} & \best{12.21} & 378.15 & 18.4 \\
      Kimi-K2.6 (1.04T/32B)~\citep{moonshot2026kimi26} & 134.30 & \runnerup{2.7} & \thirdbest{3.3} & 5.3 & 3.3 & 4.0 & 32.7 & 29.53 & 21.48 & 309.26 & 12.9 \\
      MiMo-V2.5 (310B/15B)~\citep{xiaomi2026mimov25} & 183.06 & 0.0 & 0.7 & 2.0 & 8.7 & 4.7 & 32.0 & 29.97 & 38.43 & 306.20 & 13.6 \\
      \midrule
      \multicolumn{12}{l}{\textbf{\textit{Proprietary models}}} \\
      Qwen3-VL-Plus~\citep{bai2025qwen3} & 211.50 & 0.0 & 0.0 & 0.7 & \thirdbest{19.3} & \thirdbest{17.3} & 38.7 & 36.69 & 18.87 & 282.32 & \thirdbest{19.3} \\
      Gemini-2.5-Flash~\citep{comanici2025gemini} & 151.40 & 0.7 & 2.7 & \thirdbest{6.0} & 3.3 & 1.3 & 20.7 & 19.01 & 28.71 & 346.73 & 8.2 \\
      Gemini-2.5-Pro~\citep{comanici2025gemini} & \thirdbest{113.90} & 0.7 & 1.3 & 4.0 & 8.7 & 4.7 & 24.7 & 23.38 & 28.25 & 489.59 & 11.4 \\
      GPT-5~\citep{singh2025openai} & 158.58 & \thirdbest{1.3} & 2.7 & 4.0 & 2.0 & 0.0 & 34.0 & 33.60 & 20.82 & \runnerup{236.11} & 12.4 \\
      GPT-5.4~\citep{openai2026gpt54} & 119.13 & \runnerup{2.7} & 2.7 & 3.3 & 6.0 & 0.7 & 37.3 & 36.41 & 22.06 & 356.06 & 15.3 \\
      Claude-Sonnet-4.6~\citep{anthropic2026sonnet46} & 141.72 & 0.0 & 1.3 & 4.7 & 10.7 & 6.7 & 35.3 & 34.29 & 19.80 & \thirdbest{250.34} & 15.3 \\
      Doubao-Seed-2.0-Pro~\citep{seed2026seed2} & \runnerup{66.88} & \runnerup{2.7} & \runnerup{6.7} & \runnerup{13.3} & \runnerup{27.3} & \runnerup{26.7} & \runnerup{43.3} & \thirdbest{40.48} & \thirdbest{13.39} & 446.21 & \runnerup{24.4} \\
      \midrule
      \muris & \best{19.16} & \best{66.0} & \best{75.3} & \best{82.7} & \best{28.0} & \best{27.3} & \best{44.7} & \best{44.00} & \runnerup{12.22} & \best{154.40} & \best{46.2} \\
      \bottomrule
    \end{tabular}%
  }
  \vspace{-1mm}
  \caption[MURIS-Bench-Scan averaged-split results]{\benchscan averaged-split results on HM3D. Accuracy metrics and collision ratios are percentages.}
  \label{tab:scan-average}
  \vspace{-4mm}
\end{table*}

\begin{table*}[t]
\centering
\begin{minipage}[t]{0.285\textwidth}
  \vspace{0pt}
  \centering
  \scriptsize
  \setlength{\tabcolsep}{1.5pt}
  \renewcommand{\arraystretch}{0.90}
  \begin{tabular*}{\linewidth}{@{\extracolsep{\fill}}lcccc@{}}
    \toprule
    \textbf{Training} & \shortstack{\textbf{Pose}\\L2$\downarrow$} & \shortstack{\textbf{Ground.}\\Hit$\uparrow$} & \shortstack{\textbf{Path}\\@0$\uparrow$} & \shortstack{\textbf{Med.}\\DTW$\downarrow$} \\
    \midrule
    \multicolumn{5}{@{}l}{\textit{Evaluation on \benchsyn}} \\
    Syn & 5.71 & 56.6 & \textbf{40.2} & 69.36 \\
    Scan & 31.67 & 23.8 & 33.8 & 302.23 \\
    Syn+Scan & \textbf{5.22} & \textbf{57.6} & \textbf{40.2} & \textbf{69.10} \\
    \midrule
    \multicolumn{5}{@{}l}{\textit{Evaluation on \benchscan}} \\
    Syn & 62.15 & 16.0 & 34.0 & 435.61 \\
    Scan & 33.82 & 22.7 & 43.3 & 167.63 \\
    Syn+Scan & \textbf{19.16} & \textbf{28.0} & \textbf{44.7} & \textbf{154.40} \\
    \bottomrule
  \end{tabular*}
  \captionsetup{font=footnotesize,skip=3pt}
  \captionof{table}{Source-composition and cross-domain results. Full metrics are provided in the appendix.}
  \label{tab:source-composition-main}
\end{minipage}\hfill
\begin{minipage}[t]{0.345\textwidth}
  \vspace{0pt}
  \centering
  \includegraphics[width=\linewidth]{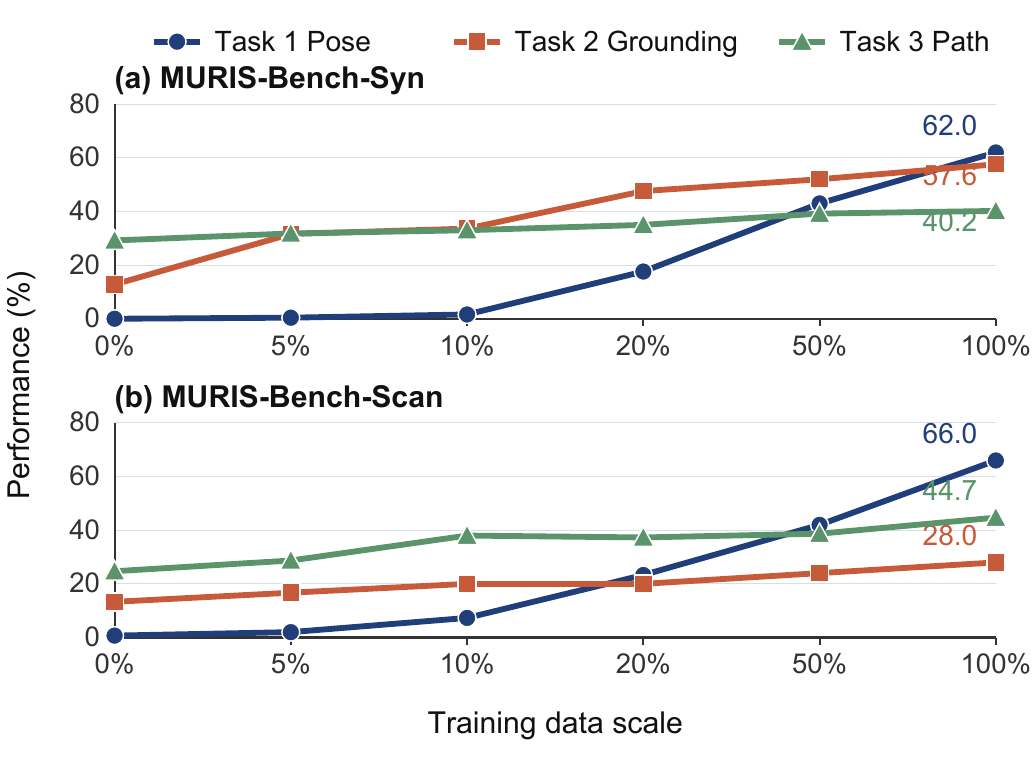}
  \vspace{-2.1mm}
  \captionsetup{font=footnotesize,skip=3pt}
  \captionof{figure}{Data-budget curves on (a) \benchsyn and (b) \benchscan.}
  \label{fig:scale}
\end{minipage}\hfill
\begin{minipage}[t]{0.345\textwidth}
  \vspace{0pt}
  \centering
  \includegraphics[width=\linewidth]{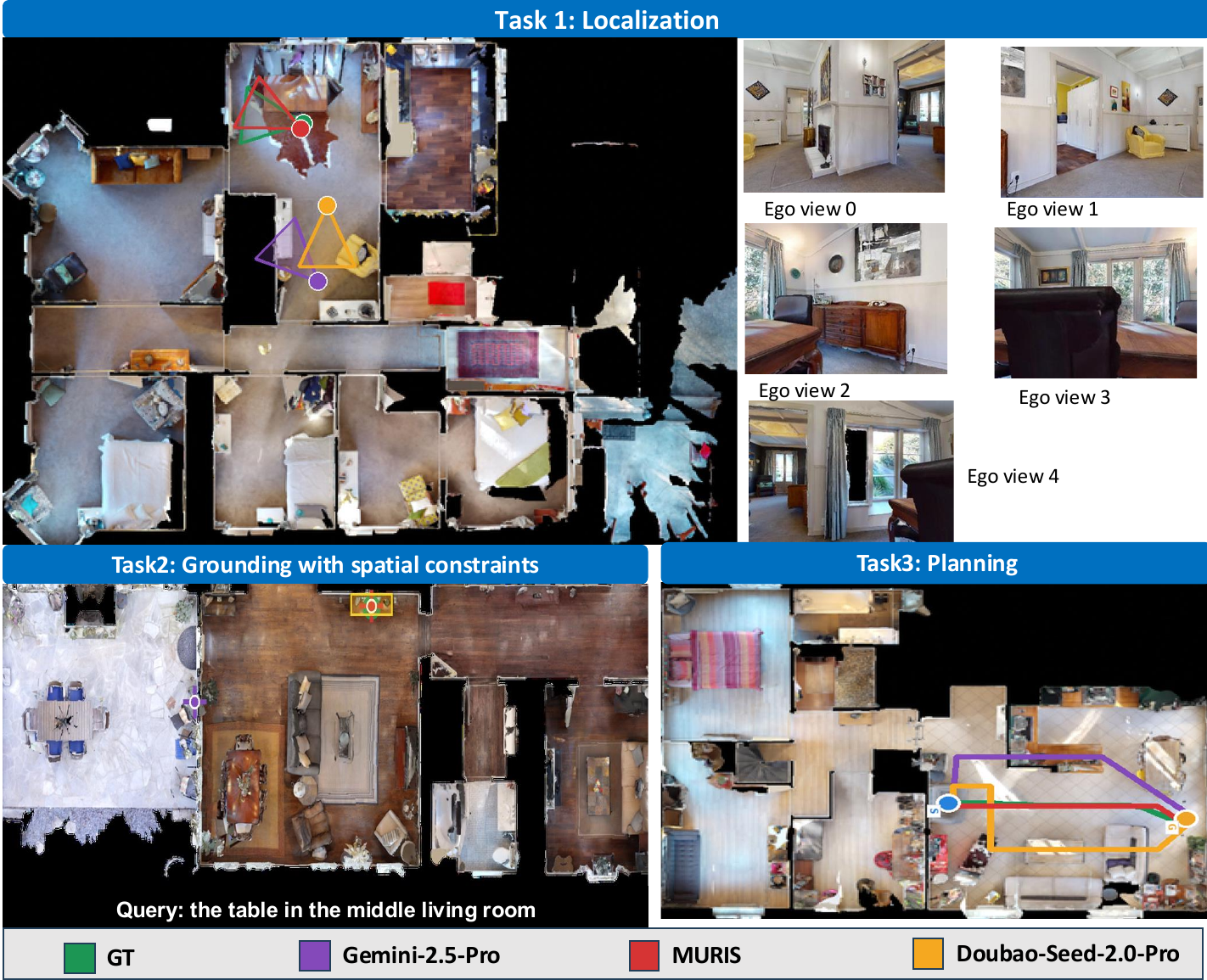}
  \vspace{-1.5mm}
  \captionsetup{font=footnotesize,skip=3pt}
  \captionof{figure}{Qualitative comparison on \benchscan across all three tasks.}
  \label{fig:case-study}
\end{minipage}
\vspace{-4mm}
\end{table*}

\subsection{Experimental Setup}
We evaluate 20 open-weight and proprietary MLLM configurations under task-equivalent prompts. Zero-shot results characterize benchmark difficulty, while data utility is measured through controlled Qwen3-VL-4B comparisons with domain-matched and mixed-source map supervision. All maps are rendered at a spatial resolution of approximately \(0.04\)~m/px; for evaluation, predicted and ground-truth coordinates are mapped from \(\Omega\) back to each environment's native map raster before computing task-specific pose, grounding, and collision-aware path metrics. Task~3 models receive an RGB map and start and goal coordinates. Collision scoring uses the paired navigability mask. 
Full evaluation and reproducibility details are provided in the appendix.

\subsection{MLLM Capability Boundaries}
Tables~\ref{tab:syn-average} and~\ref{tab:scan-average} report the averaged performance on \benchsyn and \benchscan, respectively. Their Overall Avg. equally weights the strictest pose accuracy, grounding Hit, and Path@0. Output parsing is not a major confound: parsing success ranges from 98.0\% to 100.0\% across all model--task--domain configurations, with most configurations reaching exactly 100.0\%. Detailed parsing rates are reported in Appendix Table~\ref{tab:parse-success}.

\textbf{Q1: General-purpose MLLMs exhibit coarse spatial priors but lack metric reliability.} Zero-shot results expose a sharp cross-task asymmetry: frontier models show non-trivial 2D grounding and routing priors on map-plane tasks (e.g., Doubao-Seed-2.0-Pro reaches 49.6\%/27.3\% Hit and 32.8\%/43.3\% Path@0, while GLM-4.6V reaches 37.4\%/41.3\% Path@0), yet every zero-shot model retains near-zero strict pose accuracy ($\le 0.2\%$ on \benchsyn and $\le 2.7\%$ on \benchscan). Averaging model-level Overall Avg. within each group, proprietary models outperform open-weight models by 16.1\% versus 12.1\% on \benchsyn and 15.2\% versus 11.4\% on \benchscan, and establish a higher upper envelope (27.5\% and 24.4\% for Doubao-Seed-2.0-Pro versus 18.6\% and 18.4\% for the strongest open-weight models). Scaling improves 2D grounding and routing within families such as InternVL3.5 and Molmo, though gains are non-monotonic in Qwen3-VL and model scale alone never unlocks cross-view pose registration.
Localization remains weak even when position and heading are scored separately (Appendix Table~\ref{tab:task1-decoupled}): zero-shot models reach at most 1.2\%/5.3\% position accuracy vs.\ 62.0\%/69.3\% for \muris, confirming that joint failures are not due to heading mismatch alone.

\subsection{Supervision Gains and Domain Transfer}

Having established the zero-shot boundary, we next examine whether map-specific supervision improves all three capabilities and how source composition affects cross-domain generalization with the same Qwen3-VL-4B backbone.

\textbf{Q2a: Map-specific supervision substantially improves all three map-reasoning capabilities.} Relative to the zero-shot Qwen3-VL-4B backbone, the jointly trained \muris reduces mean pose error on \benchsyn from 97.09 to 5.22 px, raises grounding Hit from 12.8\% to 57.6\%, improves Path@0 from 29.2\% to 40.2\%, raises collision-free SPL from 26.53\% to 39.17\%, and reduces collision ratio from 24.19\% to 14.99\% and median DTW from 114.48 to 69.10 px. On \benchscan, it reduces mean pose error from 185.91 to 19.16 px, raises grounding Hit from 13.3\% to 28.0\%, improves Path@0 from 24.7\% to 44.7\%, raises SPL from 23.44\% to 44.00\%, and reduces collision ratio from 47.12\% to 12.22\% and median DTW from 422.72 to 154.40 px. 
Overall Avg. rises from 14.0\% to 53.3\% on \benchsyn and from 12.7\% to 46.2\% on \benchscan, enabling the 4B \muris model to surpass all evaluated proprietary models (up to 27.5\% and 24.4\%); gains are largest in cross-view localization, whereas margins over frontier baselines remain tighter on scan-domain grounding (28.0\% vs.\ 27.3\%) and routing (44.7\% vs.\ 43.3\%).

To isolate the role of supervision source, we train synthetic-only, scan-only, and mixed-source variants with the same LoRA and optimization recipe, and evaluate every variant on both benchmark domains. This controlled comparison tests whether mixed-source supervision can produce one unified model rather than separate domain-specific models.
Table~\ref{tab:source-composition-main} reports one primary metric for localization and grounding and two complementary planning metrics; the complete metric suite is reported in the appendix.

\textbf{Q2b: Mixed-source supervision yields a unified model with higher aggregate performance than single-source training in both synthetic and scan-derived domains.}
Single-source models exhibit substantial cross-domain degradation: relative to Syn$\rightarrow$Syn, Scan$\rightarrow$Syn increases pose error from 5.71 to 31.67 px and reduces grounding Hit from 56.6\% to 23.8\%; relative to Scan$\rightarrow$Scan, Syn$\rightarrow$Scan increases pose error from 33.82 to 62.15 px and reduces grounding Hit from 22.7\% to 16.0\%. In contrast, the mixed-source model preserves \benchsyn performance while improving \benchscan over Scan-only training in pose error (33.82 to 19.16 px), grounding Hit (22.7\% to 28.0\%), Path@0 (43.3\% to 44.7\%), and median DTW (167.63 to 154.40 px). 
These results show that mixed-source training is not simply a compromise between domains. It retains the synthetic-domain strengths of Syn-only supervision while improving all four reported scan metrics over Scan-only training, most notably reducing pose error by 43.3\%. The two sources therefore provide complementary rather than interchangeable supervision, supporting \murisinstruct as a unified data engine for training a single map-reasoning checkpoint.

\input{secs/synthetic_difficulty_table.tex}

\subsection{Semantic and Geometric Capability Analysis}

\textbf{Q3: Difficulty-stratified analysis reveals distinct semantic and geometric challenges.} Aggregate metrics obscure the task-specific conditions governing post-supervision performance across layout complexity and domain realism.

\textbf{Synthetic difficulty reveals a hierarchy of map-reasoning capabilities.} As detailed in Appendix Table~\ref{tab:syn-difficulty}, \muris maintains strong localization on the \benchsyn hard split (63.0\% Acc@5/15), whereas grounding and planning are far more sensitive to layout complexity: Hit drops from 78.5\% on easy cases to 38.0\% on medium and 55.0\% on hard cases, while Path@0 drops from 73.5\% to 22.0\% and 10.0\%. Notably, relaxed Path@2 remains high on medium and hard splits (84.0\% and 62.0\%), indicating that global route topology largely holds while minor obstacle grazing in tight corridors drives strict Path@0 failures.

\textbf{Scan-derived maps provide a sharper diagnostic view of task-specific capability boundaries.}
With mixed-source supervision, \muris reaches 66.0\% Acc@15/45 for localization, 28.0\% grounding Hit, and 44.7\% Path@0 on \benchscan (Appendix Table~\ref{tab:scan-task3}).
HM3D maps combine irregular layouts, diverse object appearances, clutter, scan artifacts, larger scene footprints, and broader room semantics. Together, difficulty-stratified and route-level diagnostics isolate two dominant bottlenecks: multi-instance disambiguation for grounding and tight obstacle-clearance geometry for planning.

\subsection{Capability Scaling with Supervision}

\textbf{Q4: The three capabilities exhibit different data-scale responses.} We analyze performance across varying training-data budgets in Figure~\ref{fig:scale}. Task 1 uses Acc@5/15 on \benchsyn and Acc@15/45 on \benchscan; Task 2 uses grounding Hit; and Task 3 uses Path@0. The 0\% point represents the zero-shot backbone, and each supervised budget uses a frozen subset at the indicated supervision percentage. On \benchsyn, ego-to-map localization is the most data responsive: Task 1 Acc@5/15 rises from near zero at 5--10\% supervision to 62.0\% at full scale. Target grounding improves more gradually, while Task 3 Path@0 increases from 29.2\% zero-shot to 40.2\% at full scale. The scan-derived curves also improve unevenly, with full supervision reaching 66.0\% Acc@15/45, 28.0\% grounding Hit, and 44.7\% Path@0. These results show that additional data is especially effective for localization, whereas precise grounding and collision-free planning receive smaller returns from scaling alone.

\subsection{Qualitative Analysis}
Figure~\ref{fig:case-study} visually corroborates the quantitative failure analysis; the appendix provides additional visualizations across models and scenes. In Task 1, \muris places the pose close to the ground-truth location with the correct heading, whereas Gemini-2.5-Pro and Doubao-Seed-2.0-Pro localize the same ego views to different map regions with large orientation errors. In Task 2 (``the table in the middle living room''), \muris falls inside the target-instance box, while Gemini-2.5-Pro and Doubao-Seed-2.0-Pro predict distant points outside the target instance. In Task 3, only the \muris route satisfies the collision-free criterion; the visually plausible detours from Gemini-2.5-Pro and Doubao-Seed-2.0-Pro cross obstacles. These examples distinguish spatially plausible outputs from geometrically reliable predictions.

%% file: secs/synthetic_difficulty_table.tex
\newcommand{\syntheticdifficultytable}{%
\begin{table*}[!t]
\centering
\tiny
\setlength{\tabcolsep}{1.8pt}
\renewcommand{\arraystretch}{0.62}
\resizebox{\textwidth}{!}{%
\begin{tabular}{CCCCCCCCCCC}
\toprule
\multirow{2}{*}{Model} & \multirow{2}{*}{Split} &
\multicolumn{3}{c}{Task~1 Pose} &
\multicolumn{2}{c}{Task~2 Grounding} &
\multicolumn{3}{c}{Task~3 Path} & \multirow{2}{*}{Overall} \\
\cmidrule(lr){3-5}\cmidrule(lr){6-7}\cmidrule(lr){8-10}
& & Mean L2$\downarrow$ & Acc@5/15$\uparrow$ & Acc@10/30$\uparrow$ &
Hit$\uparrow$ & Acc@10$\uparrow$ & Path@0$\uparrow$ & Path@2$\uparrow$ & Med. DTW$\downarrow$ & Avg.$\uparrow$ \\
\midrule
\multicolumn{11}{l}{\textbf{\textit{Open-weight models}}} \\
\multirow{3}{*}{InternVL3.5-8B~\citep{wang2025internvl3} } & easy & 69.81 & 0.0 & 0.0 & 15.0 & 19.0 & 43.0 & 52.5 & 129.27 & 19.3 \\
\arrayrulecolor{black!10}\cline{2-11}\arrayrulecolor{black}
 & medium & 120.64 & \runnerup{0.5} & \runnerup{0.5} & 7.0 & 12.5 & 12.5 & 31.0 & 282.96 & 6.7 \\
\arrayrulecolor{black!10}\cline{2-11}\arrayrulecolor{black}
 & hard & 159.99 & 0.0 & 0.0 & 1.0 & 3.0 & 4.0 & 15.0 & 540.33 & 1.7 \\
\arrayrulecolor{black!18}\hline\arrayrulecolor{black}
\multirow{3}{*}{InternVL3.5-38B~\citep{wang2025internvl3}} & easy & 65.33 & 0.0 & 0.0 & \thirdbest{31.5} & 37.0 & 62.5 & 83.0 & 37.50 & 31.3 \\
\arrayrulecolor{black!10}\cline{2-11}\arrayrulecolor{black}
 & medium & 108.87 & 0.0 & 0.0 & 16.0 & 17.5 & 14.5 & 54.5 & 132.47 & 10.2 \\
\arrayrulecolor{black!10}\cline{2-11}\arrayrulecolor{black}
 & hard & 144.77 & 0.0 & 0.0 & \thirdbest{26.0} & \thirdbest{32.0} & 4.0 & 35.0 & 250.64 & \thirdbest{10.0} \\
\arrayrulecolor{black!18}\hline\arrayrulecolor{black}
\multirow{3}{*}{Molmo-7B-D} & easy & 64.73 & 0.0 & 0.0 & 10.5 & 15.0 & 7.5 & 9.0 & 799.78 & 6.0 \\
\arrayrulecolor{black!10}\cline{2-11}\arrayrulecolor{black}
 & medium & 112.18 & 0.0 & \runnerup{0.5} & 3.0 & 6.5 & 1.5 & 4.0 & 758.99 & 1.5 \\
\arrayrulecolor{black!10}\cline{2-11}\arrayrulecolor{black}
 & hard & 148.79 & 0.0 & 0.0 & 3.0 & 6.0 & 0.0 & 4.0 & 1036.53 & 1.0 \\
\arrayrulecolor{black!18}\hline\arrayrulecolor{black}
\multirow{3}{*}{Molmo-72B-0924} & easy & 69.02 & 0.0 & 0.0 & 20.5 & 33.5 & 51.5 & 70.0 & 51.60 & 24.0 \\
\arrayrulecolor{black!10}\cline{2-11}\arrayrulecolor{black}
 & medium & 114.12 & 0.0 & 0.0 & 11.5 & 23.0 & 9.5 & 51.5 & 172.67 & 7.0 \\
\arrayrulecolor{black!10}\cline{2-11}\arrayrulecolor{black}
 & hard & 150.70 & 0.0 & 0.0 & 9.0 & 13.0 & \thirdbest{7.0} & 32.0 & 296.10 & 5.3 \\
\arrayrulecolor{black!18}\hline\arrayrulecolor{black}
\multirow{3}{*}{RoboBrain2.5-8B} & easy & 71.23 & \runnerup{0.5} & 0.5 & 15.0 & 19.5 & 32.5 & 48.0 & 143.63 & 16.0 \\
\arrayrulecolor{black!10}\cline{2-11}\arrayrulecolor{black}
 & medium & 137.16 & 0.0 & \runnerup{0.5} & 8.0 & 11.0 & 9.0 & 32.5 & 220.36 & 5.7 \\
\arrayrulecolor{black!10}\cline{2-11}\arrayrulecolor{black}
 & hard & 157.82 & 0.0 & 0.0 & 10.0 & 14.0 & 4.0 & 24.0 & 323.44 & 4.7 \\
\arrayrulecolor{black!18}\hline\arrayrulecolor{black}
\multirow{3}{*}{MolmoPoint-8B} & easy & 92.44 & 0.0 & 0.0 & 6.0 & 8.5 & 58.0 & 71.0 & 87.17 & 21.3 \\
\arrayrulecolor{black!10}\cline{2-11}\arrayrulecolor{black}
 & medium & 147.54 & 0.0 & 0.0 & 0.5 & 0.5 & 16.5 & 36.5 & 260.82 & 5.7 \\
\arrayrulecolor{black!10}\cline{2-11}\arrayrulecolor{black}
 & hard & 187.49 & 0.0 & 0.0 & 1.0 & 1.0 & 5.0 & 19.0 & 366.54 & 2.0 \\
\arrayrulecolor{black!18}\hline\arrayrulecolor{black}
\multirow{3}{*}{LLaVA-OneVision-2-8B} & easy & 105.82 & 0.0 & 0.5 & 4.5 & 5.5 & 52.5 & 62.5 & 67.56 & 19.0 \\
\arrayrulecolor{black!10}\cline{2-11}\arrayrulecolor{black}
 & medium & 164.40 & 0.0 & 0.0 & 2.5 & 2.5 & 8.5 & 27.0 & 446.81 & 3.7 \\
\arrayrulecolor{black!10}\cline{2-11}\arrayrulecolor{black}
 & hard & 195.21 & 0.0 & 0.0 & 2.0 & 2.0 & 2.0 & 10.0 & 785.79 & 1.3 \\
\arrayrulecolor{black!18}\hline\arrayrulecolor{black}
\multirow{3}{*}{Qwen3-VL-4B~\citep{bai2025qwen3}} & easy & 63.85 & 0.0 & 0.0 & 22.0 & 26.5 & 58.5 & 88.5 & 41.86 & 26.8 \\
\arrayrulecolor{black!10}\cline{2-11}\arrayrulecolor{black}
 & medium & 112.77 & 0.0 & 0.0 & 7.5 & 17.0 & 13.0 & 58.0 & 139.69 & 6.8 \\
\arrayrulecolor{black!10}\cline{2-11}\arrayrulecolor{black}
 & hard & 132.18 & 0.0 & 0.0 & 5.0 & 8.0 & 3.0 & 44.0 & 244.39 & 2.7 \\
\arrayrulecolor{black!18}\hline\arrayrulecolor{black}
\multirow{3}{*}{Qwen3-VL-8B~\citep{bai2025qwen3}} & easy & 72.18 & 0.0 & 0.0 & 9.5 & 12.0 & 62.5 & 89.0 & 38.35 & 24.0 \\
\arrayrulecolor{black!10}\cline{2-11}\arrayrulecolor{black}
 & medium & 112.57 & \runnerup{0.5} & \runnerup{0.5} & 3.0 & 9.0 & 14.5 & 59.5 & 130.08 & 6.0 \\
\arrayrulecolor{black!10}\cline{2-11}\arrayrulecolor{black}
 & hard & 141.95 & 0.0 & 0.0 & 3.0 & 8.0 & 5.0 & 51.0 & 257.40 & 2.7 \\
\arrayrulecolor{black!18}\hline\arrayrulecolor{black}
\multirow{3}{*}{Qwen3-VL-32B~\citep{bai2025qwen3}} & easy & 66.60 & 0.0 & 0.0 & 25.0 & 32.5 & 64.5 & 87.5 & 34.95 & 29.8 \\
\arrayrulecolor{black!10}\cline{2-11}\arrayrulecolor{black}
 & medium & 103.17 & 0.0 & 0.0 & 8.5 & 16.5 & 9.0 & 49.5 & 148.12 & 5.8 \\
\arrayrulecolor{black!10}\cline{2-11}\arrayrulecolor{black}
 & hard & 118.31 & 0.0 & 0.0 & 2.0 & 11.0 & 3.0 & 40.0 & 248.72 & 1.7 \\
\arrayrulecolor{black!18}\hline\arrayrulecolor{black}
\multirow{3}{*}{GLM-4.6V} & easy & 65.98 & 0.0 & 0.0 & 6.5 & 12.0 & \runnerup{72.0} & \best{97.5} & \runnerup{23.12} & 26.2 \\
\arrayrulecolor{black!10}\cline{2-11}\arrayrulecolor{black}
 & medium & 116.70 & 0.0 & 0.0 & 8.0 & 15.0 & \runnerup{19.0} & \best{86.0} & \thirdbest{111.48} & 9.0 \\
\arrayrulecolor{black!10}\cline{2-11}\arrayrulecolor{black}
 & hard & 155.29 & 0.0 & 0.0 & 8.0 & 15.0 & 5.0 & \thirdbest{57.0} & 224.63 & 4.3 \\
\arrayrulecolor{black!18}\hline\arrayrulecolor{black}
\multirow{3}{*}{Kimi-K2.6} & easy & 67.01 & 0.0 & \thirdbest{1.0} & 6.5 & 8.5 & 63.0 & 86.0 & 41.50 & 23.2 \\
\arrayrulecolor{black!10}\cline{2-11}\arrayrulecolor{black}
 & medium & 100.94 & 0.0 & 0.0 & 1.0 & 8.5 & 12.5 & 44.0 & 167.12 & 4.5 \\
\arrayrulecolor{black!10}\cline{2-11}\arrayrulecolor{black}
 & hard & 106.81 & 0.0 & \runnerup{1.0} & 3.0 & 6.0 & 3.0 & 27.0 & 296.83 & 2.0 \\
\arrayrulecolor{black!18}\hline\arrayrulecolor{black}
\multirow{3}{*}{MiMo-V2.5} & easy & 67.97 & 0.0 & 0.0 & 9.0 & 12.5 & 67.0 & \runnerup{94.5} & 33.52 & 25.3 \\
\arrayrulecolor{black!10}\cline{2-11}\arrayrulecolor{black}
 & medium & 110.99 & 0.0 & \runnerup{0.5} & 3.0 & 6.0 & 17.0 & 62.0 & 127.15 & 6.7 \\
\arrayrulecolor{black!10}\cline{2-11}\arrayrulecolor{black}
 & hard & 119.55 & 0.0 & 0.0 & 0.0 & 3.0 & 4.0 & 37.0 & 266.35 & 1.3 \\
\midrule
\multicolumn{11}{l}{\textbf{\textit{Proprietary models}}} \\
\multirow{3}{*}{Qwen3-VL-Plus~\citep{bai2025qwen3}} & easy & 72.20 & 0.0 & 0.5 & 28.0 & 37.0 & \thirdbest{68.5} & 92.5 & 34.43 & \thirdbest{32.2} \\
\arrayrulecolor{black!10}\cline{2-11}\arrayrulecolor{black}
 & medium & 123.97 & 0.0 & \runnerup{0.5} & 13.5 & 22.0 & 9.5 & 60.5 & 138.24 & 7.7 \\
\arrayrulecolor{black!10}\cline{2-11}\arrayrulecolor{black}
 & hard & 154.59 & 0.0 & 0.0 & 11.0 & 17.0 & 4.0 & 35.0 & 236.50 & 5.0 \\
\arrayrulecolor{black!18}\hline\arrayrulecolor{black}
\multirow{3}{*}{Gemini-2.5-Flash~\citep{comanici2025gemini}} & easy & 57.18 & 0.0 & 0.0 & 8.0 & 13.5 & 53.5 & 73.5 & 53.73 & 20.5 \\
\arrayrulecolor{black!10}\cline{2-11}\arrayrulecolor{black}
 & medium & 73.62 & 0.0 & 0.0 & 1.5 & 5.0 & 11.0 & 36.5 & 183.54 & 4.2 \\
\arrayrulecolor{black!10}\cline{2-11}\arrayrulecolor{black}
 & hard & 105.98 & 0.0 & 0.0 & 1.0 & 3.0 & 5.0 & 16.0 & 307.10 & 2.0 \\
\arrayrulecolor{black!18}\hline\arrayrulecolor{black}
\multirow{3}{*}{Gemini-2.5-Pro~\citep{comanici2025gemini}} & easy & \thirdbest{54.00} & 0.0 & 0.5 & 29.5 & \thirdbest{41.5} & 62.0 & 93.5 & \thirdbest{31.02} & 30.5 \\
\arrayrulecolor{black!10}\cline{2-11}\arrayrulecolor{black}
 & medium & 73.29 & 0.0 & 0.0 & \thirdbest{16.5} & \thirdbest{27.0} & \thirdbest{18.0} & \thirdbest{80.0} & \runnerup{109.82} & \thirdbest{11.5} \\
\arrayrulecolor{black!10}\cline{2-11}\arrayrulecolor{black}
 & hard & 95.17 & 0.0 & 0.0 & 10.0 & 20.0 & 2.0 & \best{70.0} & \runnerup{205.25} & 4.0 \\
\arrayrulecolor{black!18}\hline\arrayrulecolor{black}
\multirow{3}{*}{GPT-5~\citep{singh2025openai} } & easy & 57.74 & 0.0 & \thirdbest{1.0} & 8.5 & 11.5 & 66.0 & 80.0 & 48.31 & 24.8 \\
\arrayrulecolor{black!10}\cline{2-11}\arrayrulecolor{black}
 & medium & \thirdbest{69.53} & 0.0 & \runnerup{0.5} & 0.5 & 3.0 & 11.5 & 49.5 & 192.73 & 4.0 \\
\arrayrulecolor{black!10}\cline{2-11}\arrayrulecolor{black}
 & hard & \thirdbest{95.01} & 0.0 & 0.0 & 0.0 & 1.0 & \runnerup{8.0} & 31.0 & 323.18 & 2.7 \\
\arrayrulecolor{black!18}\hline\arrayrulecolor{black}
\multirow{3}{*}{GPT-5.4} & easy & 59.60 & \runnerup{0.5} & \runnerup{2.5} & 15.5 & 22.0 & 64.5 & 92.5 & 36.94 & 26.8 \\
\arrayrulecolor{black!10}\cline{2-11}\arrayrulecolor{black}
 & medium & 88.02 & 0.0 & 0.0 & 2.5 & 4.5 & \thirdbest{18.0} & 65.5 & 128.12 & 6.8 \\
\arrayrulecolor{black!10}\cline{2-11}\arrayrulecolor{black}
 & hard & 114.91 & 0.0 & 0.0 & 2.0 & 8.0 & 5.0 & 46.0 & \thirdbest{208.57} & 2.3 \\
\arrayrulecolor{black!18}\hline\arrayrulecolor{black}
\multirow{3}{*}{Claude-Sonnet-4.6} & easy & 63.67 & 0.0 & 0.5 & 23.0 & 27.5 & 65.5 & 88.0 & 38.28 & 29.5 \\
\arrayrulecolor{black!10}\cline{2-11}\arrayrulecolor{black}
 & medium & 97.20 & 0.0 & \runnerup{0.5} & 8.5 & 15.5 & 11.5 & 46.0 & 150.79 & 6.7 \\
\arrayrulecolor{black!10}\cline{2-11}\arrayrulecolor{black}
 & hard & 136.38 & 0.0 & 0.0 & 1.0 & 8.0 & 2.0 & 27.0 & 275.71 & 1.0 \\
\arrayrulecolor{black!18}\hline\arrayrulecolor{black}
\multirow{3}{*}{Doubao-Seed-2.0-Pro} & easy & \runnerup{53.67} & 0.0 & \thirdbest{1.0} & \runnerup{48.0} & \runnerup{50.0} & 59.0 & 92.0 & 40.85 & \runnerup{35.7} \\
\arrayrulecolor{black!10}\cline{2-11}\arrayrulecolor{black}
 & medium & \runnerup{61.43} & 0.0 & 0.0 & \best{53.5} & \best{57.5} & \thirdbest{18.0} & 71.5 & 116.24 & \runnerup{23.8} \\
\arrayrulecolor{black!10}\cline{2-11}\arrayrulecolor{black}
 & hard & \runnerup{70.48} & 0.0 & \runnerup{1.0} & \runnerup{45.0} & \runnerup{49.0} & \best{10.0} & 55.0 & 221.13 & \runnerup{18.3} \\
\midrule
\multicolumn{11}{l}{\textbf{\textit{Data-trained map specialist}}} \\
\multirow{3}{*}{\muris} & easy & \best{4.24} & \best{73.5} & \best{96.0} & \best{78.5} & \best{82.0} & \best{73.5} & \thirdbest{94.0} & \best{12.46} & \best{75.2} \\
\arrayrulecolor{black!10}\cline{2-11}\arrayrulecolor{black}
 & medium & \best{5.96} & \best{50.0} & \best{87.0} & \runnerup{38.0} & \runnerup{49.5} & \best{22.0} & \runnerup{84.0} & \best{92.38} & \best{36.7} \\
\arrayrulecolor{black!10}\cline{2-11}\arrayrulecolor{black}
 & hard & \best{5.67} & \best{63.0} & \best{95.0} & \best{55.0} & \best{66.0} & \best{10.0} & \runnerup{62.0} & \best{186.02} & \best{42.7} \\
\arrayrulecolor{black!18}\hline\arrayrulecolor{black}
\bottomrule
\end{tabular}
}
\caption[MURIS-Bench-Syn detailed difficulty results]{\benchsyn detailed difficulty results. Zero-shot models are grouped by weight availability, with the jointly trained \muris specialist shown separately. Each model occupies three rows for easy, medium, and hard splits. Overall Avg. is the arithmetic mean of Acc@5/15, grounding Hit, and Path@0. Accuracy-style metrics are percentages. Bold, underlined, and italicized values mark the best, second-best, and third-best non-zero results, with ties marked equally.}
\label{tab:syn-difficulty}
\end{table*}
}

%% file: secs/conclusion.tex
\section{Conclusion}
We introduce \murisbench to evaluate ego-to-map localization, language target grounding, and path planning directly over top-down visual maps with continuous-coordinate and collision-aware metrics,
together with a scalable data engine that produces \murisinstruct with geometrically verified examples. Across 20 MLLM configurations, general-purpose models exhibit coarse semantic and
routing priors but lack the metric precision required for reliable localization, grounding, and collision-free execution. 
Controlled same-backbone experiments show that map-specific supervision substantially improves all three capabilities without specialized architecture, while mixed-source training yields a unified model with stronger aggregate performance across both domains. Difficulty-stratified and data-budget analyses reveal capability-specific learning patterns, with instance ambiguity and obstacle-constrained geometry remaining the main challenges for grounding and planning.

%% file: secs/appendix.tex
\raggedbottom

\section{Appendix Overview}

The appendix is organized as follows:
\begin{itemize}[leftmargin=*,topsep=2pt,itemsep=0pt,parsep=0pt]
\item \textit{Additional Results}: detailed synthetic and scan results, decoupled localization analysis, mixed-source and cross-domain controls, and qualitative comparisons.
\item \textit{Additional Experimental Details}: frozen training and inference settings, prompts, parsing, aggregation, and collision scoring.
\item \textit{Data Details}: source-specific train/evaluation distributions followed by representative synthetic and scan examples.
\end{itemize}

\section{Additional Results}

\subsection{Detailed Synthetic Results}
\label{app:syn-results}

Table~\ref{tab:syn-difficulty} reports the detailed \benchsyn results across easy, medium, and hard difficulty levels, including Path@2 as a relaxed collision-threshold diagnostic. Its \muris rows report results from the same unified model as the main paper.

\syntheticdifficultytable

\subsection{Additional Scan Results}
\label{app:scan-results}

Table~\ref{tab:scan-task3} reports the detailed \benchscan Task~3 path-planning results across difficulty levels, including Path@2 as a relaxed collision-threshold diagnostic. Its \muris rows report results from the same unified model as the main paper.

\subsection{Decoupled Localization Analysis}
\label{app:task1-decoupled}

Because the primary Task~1 metrics require both accurate camera location and accurate heading, Table~\ref{tab:task1-decoupled} decomposes ego-to-map localization into separate position and orientation components across all evaluated configurations ($5\,\mathrm{px}$ and $15^{\circ}$ on \benchsyn; $15\,\mathrm{px}$ and $45^{\circ}$ on \benchscan). Under uniform random heading guesses in $[0,360)$, expected circular error is $90^{\circ}$ and thresholded heading rates are $8.3\%$ at $15^{\circ}$ and $25.0\%$ at $45^{\circ}$.

Across all zero-shot baselines, scoring position alone still yields at most $1.2\%$ Pos@5 on \benchsyn and $5.3\%$ Pos@15 on \benchscan, demonstrating that near-zero joint pose accuracy cannot be attributed to heading convention mismatch alone. Most zero-shot models also remain close to random angular error ($81.51^{\circ}$--$103.67^{\circ}$), while GPT-5.4 and Doubao-Seed-2.0-Pro recover partial coarse orientation ($65.48^{\circ}$ and $20.2\%$ Head@15 on \benchsyn; $69.65^{\circ}$/$69.26^{\circ}$ and $43.3\%$/$40.0\%$ Head@45 on \benchscan) without achieving reliable metric position. With map-specific supervision on the same Qwen3-VL-4B backbone, \muris substantially improves both components, raising Pos@5 from $0.2\%$ to $62.0\%$ and Head@15 from $9.2\%$ to $97.4\%$ ($4.08^{\circ}$ mean error) on \benchsyn, and raising Pos@15 from $1.3\%$ to $69.3\%$ and Head@45 from $27.3\%$ to $85.3\%$ ($19.99^{\circ}$ mean error) on \benchscan.

\begin{table*}[p]
\begingroup
\centering
  \scriptsize
  \setlength{\tabcolsep}{3pt}
  \renewcommand{\arraystretch}{0.82}
  \resizebox{\textwidth}{!}{%
    \begin{tabular}{CCCCCCCCCC}
      \toprule
      \multirow{2}{*}{Model} &
      \multicolumn{3}{c}{Easy} &
      \multicolumn{3}{c}{Medium} &
      \multicolumn{3}{c}{Average} \\
      \cmidrule(lr){2-4}\cmidrule(lr){5-7}\cmidrule(lr){8-10}
      & Path@0$\uparrow$ & Path@2$\uparrow$ & Med. DTW$\downarrow$ &
      Path@0$\uparrow$ & Path@2$\uparrow$ & Med. DTW$\downarrow$ &
      Path@0$\uparrow$ & Path@2$\uparrow$ & Med. DTW$\downarrow$ \\
      \midrule
      \multicolumn{10}{l}{\textbf{\textit{Open-weight models}}} \\
      InternVL3.5-8B~\citep{wang2025internvl3} & 23.3 & 50.0 & 237.97 & 19.2 & 35.8 & 489.85 & 20.0 & 38.7 & 431.14 \\
      InternVL3.5-38B~\citep{wang2025internvl3} & 43.3 & 73.3 & 143.88 & 29.2 & 52.5 & 300.60 & 32.0 & 56.7 & 266.66 \\
      Molmo-7B-D & 0.0 & 0.0 & 3882.37 & 0.8 & 3.3 & 2043.83 & 0.7 & 2.7 & 2115.17 \\
      Molmo-72B-0924 & 20.0 & 50.0 & 184.77 & 20.8 & 42.5 & 473.88 & 20.7 & 44.0 & 414.94 \\
      RoboBrain2.5-8B & 16.7 & 23.3 & 324.96 & 21.7 & 30.8 & \thirdbest{279.21} & 20.7 & 29.3 & 289.56 \\
      MolmoPoint-8B & 33.3 & 46.7 & 173.67 & 20.0 & 33.3 & 400.54 & 22.7 & 36.0 & 378.67 \\
      LLaVA-OneVision-2-8B & 30.0 & 46.7 & 139.16 & 12.5 & 27.5 & 481.69 & 16.0 & 31.3 & 402.58 \\
      Qwen3-VL-4B~\citep{bai2025qwen3} & 20.0 & 63.3 & 185.46 & 25.8 & 48.3 & 485.01 & 24.7 & 51.3 & 422.72 \\
      Qwen3-VL-8B~\citep{bai2025qwen3} & 36.7 & 76.7 & 151.77 & 28.3 & 60.0 & 417.12 & 30.0 & 63.3 & 359.80 \\
      Qwen3-VL-32B~\citep{bai2025qwen3} & \thirdbest{46.7} & 76.7 & 133.49 & 20.8 & 51.7 & 387.16 & 26.0 & 56.7 & 334.77 \\
      GLM-4.6V & \thirdbest{46.7} & \best{100.0} & 201.42 & \thirdbest{40.0} & \best{89.2} & 451.31 & \thirdbest{41.3} & \best{91.3} & 378.15 \\
      Kimi-K2.6 & \thirdbest{46.7} & \thirdbest{86.7} & 150.78 & 29.2 & 52.5 & 370.72 & 32.7 & 59.3 & 309.26 \\
      MiMo-V2.5 & 40.0 & 70.0 & 149.33 & 30.0 & 61.7 & 366.19 & 32.0 & 63.3 & 306.20 \\
      \midrule
      \multicolumn{10}{l}{\textbf{\textit{Proprietary models}}} \\
      Qwen3-VL-Plus~\citep{bai2025qwen3} & 43.3 & \runnerup{96.7} & \thirdbest{115.34} & 37.5 & 65.0 & 322.58 & 38.7 & 71.3 & 282.32 \\
      Gemini-2.5-Flash~\citep{comanici2025gemini} & 30.0 & 60.0 & 209.21 & 18.3 & 33.3 & 411.64 & 20.7 & 38.7 & 346.73 \\
      Gemini-2.5-Pro~\citep{comanici2025gemini} & 30.0 & 83.3 & 197.96 & 23.3 & \thirdbest{70.0} & 533.87 & 24.7 & \thirdbest{72.7} & 489.59 \\
      GPT-5~\citep{singh2025openai} & 36.7 & 73.3 & \runnerup{100.32} & 33.3 & 50.0 & \runnerup{276.44} & 34.0 & 54.7 & \runnerup{236.11} \\
      GPT-5.4 & \runnerup{50.0} & \thirdbest{86.7} & 132.65 & 34.2 & 60.0 & 427.09 & 37.3 & 65.3 & 356.06 \\
      Claude-Sonnet-4.6 & \runnerup{50.0} & 66.7 & 140.84 & 31.7 & 56.7 & 308.94 & 35.3 & 58.7 & \thirdbest{250.34} \\
      Doubao-Seed-2.0-Pro & 33.3 & 80.0 & 198.62 & \best{45.8} & \runnerup{84.2} & 498.46 & \runnerup{43.3} & \runnerup{83.3} & 446.21 \\
      \midrule
      \muris & \best{53.3} & 76.7 & \best{88.20} & \runnerup{42.5} & 54.2 & \best{174.04} & \best{44.7} & 58.7 & \best{154.40} \\
      \bottomrule
    \end{tabular}%
  }
  \caption[MURIS-Bench-Scan Task~3 path-planning results]{\benchscan Task~3 path-planning results on HM3D. Zero-shot models are grouped by weight availability, with the unified \muris model shown separately. Easy has 30 examples, medium has 120 examples, and the averaged split has 150 examples. Accuracy-style metrics are percentages. Bold, underlined, and italicized values mark the best, second-best, and third-best non-zero results, with ties marked equally.}
  \label{tab:scan-task3}
\endgroup
\par\vspace{5pt}
\begingroup
\centering
\scriptsize
\setlength{\tabcolsep}{2.8pt}
\renewcommand{\arraystretch}{0.82}
\resizebox{\textwidth}{!}{%
\begin{tabular}{@{}lrrrrrrrrrr@{}}
\toprule
\multirow{2}{*}{Model} & \multicolumn{5}{c}{\benchsyn} & \multicolumn{5}{c}{\benchscan} \\
\cmidrule(lr){2-6}\cmidrule(lr){7-11}
& Mean L2$\downarrow$ & Mean $\Delta\theta\downarrow$ & Pos@5$\uparrow$ & Head@15$\uparrow$ & Acc@5/15$\uparrow$
& Mean L2$\downarrow$ & Mean $\Delta\theta\downarrow$ & Pos@15$\uparrow$ & Head@45$\uparrow$ & Acc@15/45$\uparrow$ \\
\midrule
\multicolumn{11}{l}{\textbf{\textit{Open-weight models}}} \\
InternVL3.5-8B~\citep{wang2025internvl3} & 108.18 & 93.13 & \thirdbest{0.6} & 8.0 & \runnerup{0.2} & 218.23 & 89.01 & 0.0 & 30.0 & 0.0 \\
InternVL3.5-38B~\citep{wang2025internvl3} & 98.63 & 89.57 & 0.2 & 6.6 & 0.0 & 184.15 & 82.65 & 0.0 & 24.0 & 0.0 \\
Molmo-7B-D~\citep{deitke2025molmo} & 100.52 & 89.92 & 0.2 & 7.6 & 0.0 & 190.94 & 103.67 & 0.0 & 15.3 & 0.0 \\
Molmo-72B-0924~\citep{deitke2025molmo} & 103.40 & 93.26 & 0.0 & 8.8 & 0.0 & 192.84 & 102.53 & 0.0 & 14.7 & 0.0 \\
RoboBrain2.5-8B~\citep{tan2026robobrain} & 114.92 & 91.12 & 0.2 & 9.4 & \runnerup{0.2} & 224.54 & 84.27 & 0.7 & 25.3 & 0.0 \\
MolmoPoint-8B~\citep{clark2026molmopoint} & 133.49 & 90.79 & 0.2 & 8.2 & 0.0 & 242.08 & 85.93 & 0.7 & 35.3 & 0.7 \\
LLaVA-OneVision-2-8B~\citep{an2026llava} & 147.13 & 92.78 & 0.0 & 9.0 & 0.0 & 172.43 & 94.49 & 0.7 & 18.0 & 0.0 \\
Qwen3-VL-4B~\citep{bai2025qwen3} & 97.09 & 91.31 & 0.2 & 9.2 & 0.0 & 185.91 & 88.03 & 1.3 & 27.3 & 0.0 \\
Qwen3-VL-8B~\citep{bai2025qwen3} & 102.29 & 92.52 & 0.2 & 8.8 & \runnerup{0.2} & 169.34 & 89.99 & 1.3 & 32.7 & 0.0 \\
Qwen3-VL-32B~\citep{bai2025qwen3} & 91.57 & 92.20 & 0.2 & 8.2 & 0.0 & 146.67 & 83.17 & 2.0 & 36.7 & 0.7 \\
GLM-4.6V~\citep{hong2025glm} & 104.13 & 91.14 & 0.2 & 6.4 & 0.0 & 226.20 & 93.91 & 0.7 & 20.0 & 0.7 \\
Kimi-K2.6~\citep{moonshot2026kimi26} & 88.54 & 88.40 & 0.4 & 11.2 & 0.0 & 134.30 & 84.17 & 2.7 & 34.0 & \runnerup{2.7} \\
MiMo-V2.5~\citep{xiaomi2026mimov25} & 95.49 & 88.27 & 0.4 & 8.2 & 0.0 & 183.06 & 83.51 & 0.0 & 26.7 & 0.0 \\
\midrule
\multicolumn{11}{l}{\textbf{\textit{Proprietary models}}} \\
Qwen3-VL-Plus~\citep{bai2025qwen3} & 109.38 & 92.49 & 0.4 & 8.6 & 0.0 & 211.50 & 82.66 & 1.3 & 36.0 & 0.0 \\
Gemini-2.5-Flash~\citep{comanici2025gemini} & 73.51 & 91.03 & 0.4 & 9.0 & 0.0 & 151.40 & 84.41 & 2.0 & 34.7 & 0.7 \\
Gemini-2.5-Pro~\citep{comanici2025gemini} & 69.95 & 85.73 & 0.4 & 11.2 & 0.0 & \thirdbest{113.90} & 93.65 & \thirdbest{4.7} & 18.7 & 0.7 \\
GPT-5~\citep{singh2025openai} & \thirdbest{69.91} & \thirdbest{81.51} & \runnerup{1.2} & \thirdbest{11.8} & 0.0 & 158.58 & 90.88 & 1.3 & 24.0 & \thirdbest{1.3} \\
GPT-5.4~\citep{openai2026gpt54} & 82.03 & \runnerup{65.48} & 0.4 & \runnerup{20.2} & \runnerup{0.2} & 119.13 & \thirdbest{69.65} & 2.7 & \runnerup{43.3} & \runnerup{2.7} \\
Claude-Sonnet-4.6~\citep{anthropic2026sonnet46} & 91.62 & 87.64 & 0.0 & 10.4 & 0.0 & 141.72 & 84.82 & 2.0 & 31.3 & 0.0 \\
Doubao-Seed-2.0-Pro~\citep{seed2026seed2} & \runnerup{60.13} & 82.90 & \runnerup{1.2} & 10.6 & 0.0 & \runnerup{66.88} & \runnerup{69.26} & \runnerup{5.3} & \thirdbest{40.0} & \runnerup{2.7} \\
\midrule
\muris & \best{5.22} & \best{4.08} & \best{62.0} & \best{97.4} & \best{62.0} & \best{19.16} & \best{19.99} & \best{69.3} & \best{85.3} & \best{66.0} \\
\bottomrule
\end{tabular}%
}
\caption{Decoupled Task~1 position and heading results on both benchmark domains. Mean L2 is in native-map pixels and Mean $\Delta\theta$ is circular heading error in degrees. Pos@$r$ and Head@$\tau$ score position ($\le r\,\mathrm{px}$) and heading ($\le \tau^{\circ}$) independently, while Acc@$r/\tau$ requires both simultaneously. Accuracy metrics are percentages; bold, underlined, and italic values mark the top three non-zero results.}
\label{tab:task1-decoupled}
\endgroup
\end{table*}

\subsection{Mixed-Source and Cross-Domain Results}
\label{app:source-composition}

Table~\ref{tab:source-composition} evaluates the synthetic-only, scan-only, and mixed-source variants on both frozen benchmark domains. The source-specific fine-tuning prompts are used directly; no zero-shot wrapper or test-time prompt adaptation is introduced. Task~1 retains the domain-specific criteria used in the main tables: Acc@5/15 and Acc@10/30 on \benchsyn, and the relaxed Acc@15/45 and Acc@20/60 criteria on the more challenging \benchscan maps.

Mixed-source training preserves synthetic performance: relative to Syn-only training, pose error decreases from 5.71 to 5.22 px, grounding Hit increases from 56.6\% to 57.6\%, median DTW decreases from 69.36 to 69.10 px, and Overall Avg. increases from 52.3\% to 53.3\%. On \benchscan, it improves over Scan-only training in pose error (33.82 to 19.16 px), grounding Hit (22.7\% to 28.0\%), Path@0 (43.3\% to 44.7\%), median DTW (167.63 to 154.40 px), and Overall Avg. (24.2\% to 46.2\%). Under the relaxed scan-localization criteria, the mixed model reaches 66.0\% and 75.3\% at Acc@15/45 and Acc@20/60, respectively. Single-source training transfers unevenly across domains in Tasks~1 and~2, whereas mixed-source supervision achieves the highest Overall Avg. on both \benchsyn and \benchscan. The contrast between the high Scan Path@2 of Syn-only training and its lower Path@0 and higher median DTW further demonstrates that collision tolerance and route fidelity capture complementary aspects of planning quality.

\begin{table*}[p]
\begingroup
\centering
\scriptsize
\setlength{\tabcolsep}{2.6pt}
\renewcommand{\arraystretch}{0.92}
\resizebox{\textwidth}{!}{%
\begin{tabular}{CCCCCCCCCCC}
\toprule
\multicolumn{11}{l}{\textbf{Evaluation on \benchsyn}} \\
\midrule
\multirow{2}{*}{Training source} & \multirow{2}{*}{Evaluation source} &
\multicolumn{3}{c}{Task~1 Pose} &
\multicolumn{2}{c}{Task~2 Grounding} &
\multicolumn{3}{c}{Task~3 Path} & \multirow{2}{*}{Overall} \\
\cmidrule(lr){3-5}\cmidrule(lr){6-7}\cmidrule(lr){8-10}
& & Mean L2$\downarrow$ & Acc@5/15$\uparrow$ & Acc@10/30$\uparrow$ &
Hit$\uparrow$ & Acc@10$\uparrow$ & Path@0$\uparrow$ & Path@2$\uparrow$ & Med. DTW$\downarrow$ & Avg.$\uparrow$ \\
\midrule
Syn & Syn & 5.71 & 60.0 & 91.6 & 56.6 & \textbf{65.8} & \textbf{40.2} & \textbf{84.6} & 69.36 & 52.3 \\
Scan & Syn & 31.67 & 0.2 & 2.6 & 23.8 & 30.2 & 33.8 & 47.4 & 302.23 & 19.3 \\
Syn+Scan & Syn & \textbf{5.22} & \textbf{62.0} & \textbf{92.2} & \textbf{57.6} & \textbf{65.8} & \textbf{40.2} & 83.6 & \textbf{69.10} & \textbf{53.3} \\
\bottomrule
\end{tabular}%
}

\vspace{1.5mm}

\resizebox{\textwidth}{!}{%
\begin{tabular}{CCCCCCCCCCC}
\toprule
\multicolumn{11}{l}{\textbf{Evaluation on \benchscan}} \\
\midrule
\multirow{2}{*}{Training source} & \multirow{2}{*}{Evaluation source} &
\multicolumn{3}{c}{Task~1 Pose} &
\multicolumn{2}{c}{Task~2 Grounding} &
\multicolumn{3}{c}{Task~3 Path} & \multirow{2}{*}{Overall} \\
\cmidrule(lr){3-5}\cmidrule(lr){6-7}\cmidrule(lr){8-10}
& & Mean L2$\downarrow$ & Acc@15/45$\uparrow$ & Acc@20/60$\uparrow$ &
Hit$\uparrow$ & Acc@10$\uparrow$ & Path@0$\uparrow$ & Path@2$\uparrow$ & Med. DTW$\downarrow$ & Avg.$\uparrow$ \\
\midrule
Syn & Scan & 62.15 & 13.3 & 22.0 & 16.0 & 14.7 & 34.0 & \textbf{89.3} & 435.61 & 21.1 \\
Scan & Scan & 33.82 & 6.7 & 13.3 & 22.7 & 24.0 & 43.3 & 54.7 & 167.63 & 24.2 \\
Syn+Scan & Scan & \textbf{19.16} & \textbf{66.0} & \textbf{75.3} & \textbf{28.0} & \textbf{27.3} & \textbf{44.7} & 58.7 & \textbf{154.40} & \textbf{46.2} \\
\bottomrule
\end{tabular}%
}
\caption{Controlled source-composition and cross-domain evaluation. All variants use the configuration in Table~\ref{tab:training-config} and the metrics from the main benchmark tables. Syn+Scan reproduces the primary \muris results. Bold marks the best values within each evaluation domain, including ties.}
\label{tab:source-composition}
\endgroup
\par\vspace{5pt}
\begingroup
\centering
\scriptsize
\setlength{\tabcolsep}{5.2pt}
\renewcommand{\arraystretch}{0.92}
\begin{tabular*}{\textwidth}{@{\extracolsep{\fill}}lrrrrrr@{}}
\toprule
\multirow{2}{*}{Model} & \multicolumn{3}{c}{\benchsyn} & \multicolumn{3}{c}{\benchscan} \\
\cmidrule(lr){2-4}\cmidrule(lr){5-7}
& Task 1 & Task 2 & Task 3 & Task 1 & Task 2 & Task 3 \\
\midrule
InternVL3.5-8B & 100.0 & 100.0 & 100.0 & 100.0 & 100.0 & 100.0 \\
InternVL3.5-38B & 100.0 & 99.8 & 100.0 & 100.0 & 100.0 & 100.0 \\
Molmo-7B-D & 100.0 & 100.0 & 100.0 & 100.0 & 100.0 & 100.0 \\
Molmo-72B-0924 & 100.0 & 100.0 & 100.0 & 100.0 & 100.0 & 100.0 \\
RoboBrain2.5-8B & 100.0 & 100.0 & 100.0 & 100.0 & 100.0 & 100.0 \\
MolmoPoint-8B & 100.0 & 100.0 & 100.0 & 100.0 & 100.0 & 100.0 \\
LLaVA-OneVision-2-8B & 100.0 & 100.0 & 100.0 & 100.0 & 100.0 & 100.0 \\
Qwen3-VL-4B & 100.0 & 100.0 & 100.0 & 100.0 & 100.0 & 100.0 \\
Qwen3-VL-8B & 100.0 & 100.0 & 100.0 & 100.0 & 100.0 & 100.0 \\
Qwen3-VL-32B & 100.0 & 100.0 & 100.0 & 100.0 & 100.0 & 100.0 \\
GLM-4.6V & 100.0 & 100.0 & 100.0 & 100.0 & 100.0 & 100.0 \\
Kimi-K2.6 & 100.0 & 99.0 & 100.0 & 100.0 & 98.0 & 100.0 \\
MiMo-V2.5 & 100.0 & 100.0 & 100.0 & 100.0 & 100.0 & 100.0 \\
Qwen3-VL-Plus & 100.0 & 100.0 & 100.0 & 100.0 & 100.0 & 100.0 \\
Gemini-2.5-Flash & 100.0 & 99.2 & 100.0 & 100.0 & 100.0 & 100.0 \\
Gemini-2.5-Pro & 100.0 & 100.0 & 100.0 & 100.0 & 100.0 & 100.0 \\
GPT-5 & 100.0 & 100.0 & 100.0 & 100.0 & 100.0 & 100.0 \\
GPT-5.4 & 100.0 & 100.0 & 100.0 & 100.0 & 100.0 & 100.0 \\
Claude-Sonnet-4.6 & 99.8 & 99.8 & 100.0 & 100.0 & 100.0 & 100.0 \\
Doubao-Seed-2.0-Pro & 100.0 & 100.0 & 100.0 & 100.0 & 100.0 & 100.0 \\
\muris & 99.8 & 100.0 & 100.0 & 100.0 & 100.0 & 100.0 \\
\bottomrule
\end{tabular*}
\caption{Output parsing success rates (\%). Task~1 requires a valid location--orientation tuple, Task~2 a valid coordinate pair, and Task~3 at least two valid waypoint coordinates. Rates use all examples in each frozen benchmark split as the denominator.}
\label{tab:parse-success}
\endgroup
\end{table*}

\twocolumn[{%
\begin{minipage}{\textwidth}
\subsection{Additional Qualitative Results}
\label{app:qualitative}

Figures~\ref{fig:qualitative-scan-intern-glm}--\ref{fig:qualitative-syn-gemini-doubao} compare \muris with representative open-weight and proprietary models on scan-derived and synthetic maps. Green denotes ground truth, red denotes \muris, and purple and orange denote the two baselines named in each figure.

\centering
\begin{minipage}[t]{0.47\textwidth}
\centering
\includegraphics[width=0.74\linewidth]{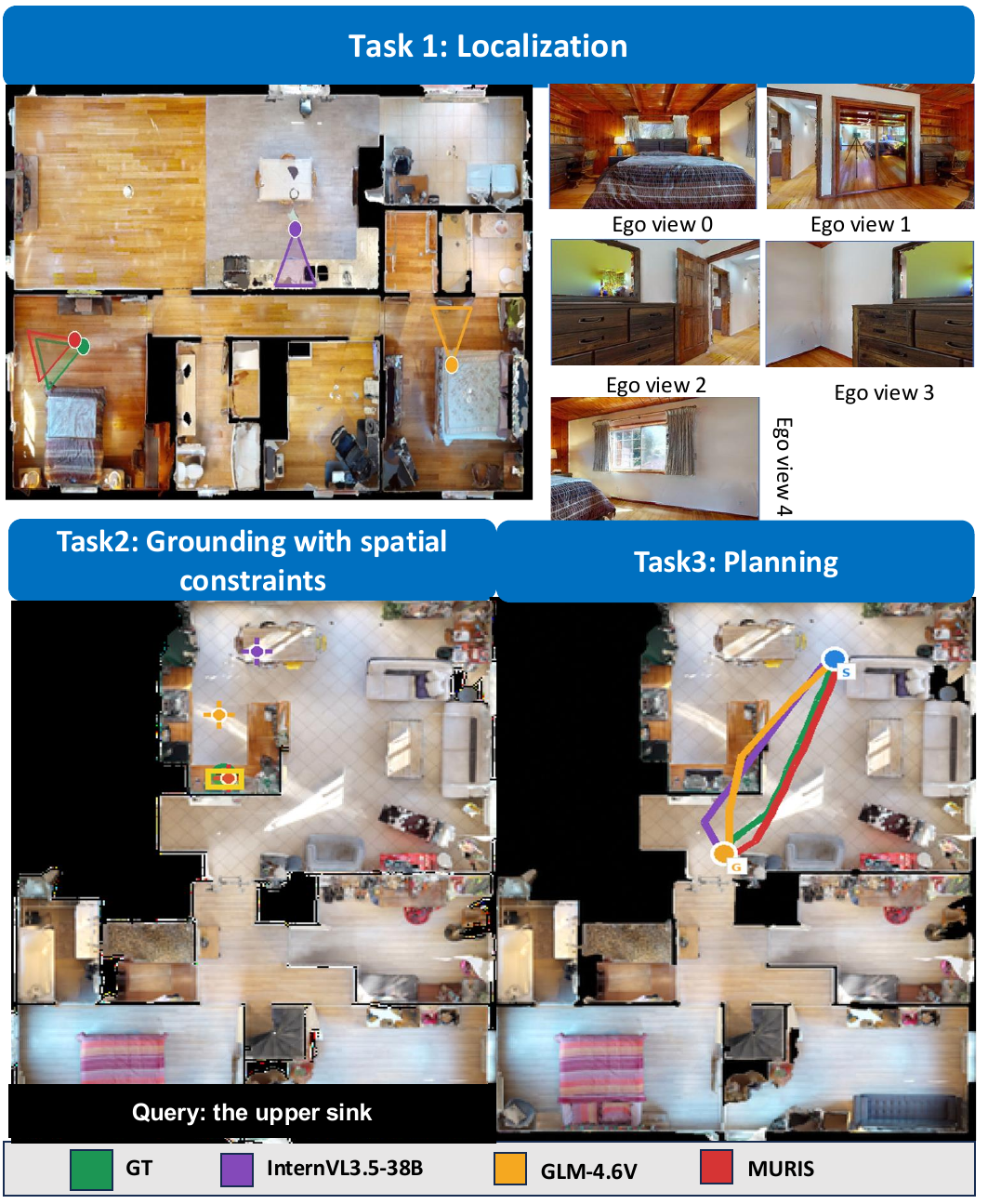}
\captionof{figure}{\textbf{Scan-derived comparison with InternVL3.5-38B and GLM-4.6V.} The Task~2 query is ``the upper sink.'' Both baselines mislocalize the Task~1 pose and select non-target regions for Task~2. Their Task~3 paths also deviate more strongly from the reference route, whereas \muris remains close to the ground truth across all three tasks.}
\label{fig:qualitative-scan-intern-glm}

\vspace{1mm}

\includegraphics[width=0.74\linewidth]{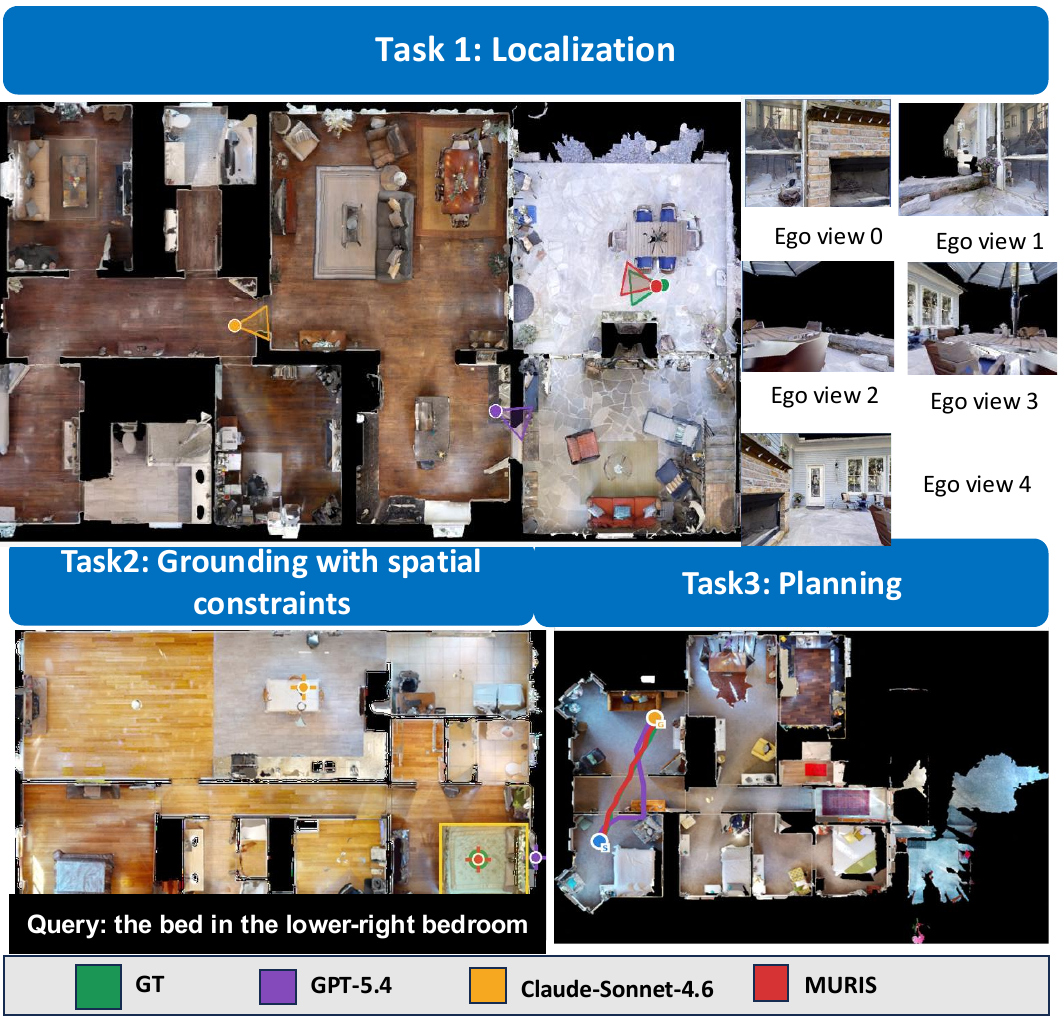}
\captionof{figure}{\textbf{Scan-derived comparison with GPT-5.4 and Claude-Sonnet-4.6.} The Task~2 query is ``the bed in the lower-right bedroom.'' The baselines place the Task~1 pose in different rooms and ground Task~2 away from the specified bed. Their planned routes exhibit larger geometric deviation, while \muris follows the reference more closely.}
\label{fig:qualitative-scan-gpt-claude}
\end{minipage}\hfill
\begin{minipage}[t]{0.47\textwidth}
\centering
\includegraphics[width=0.74\linewidth]{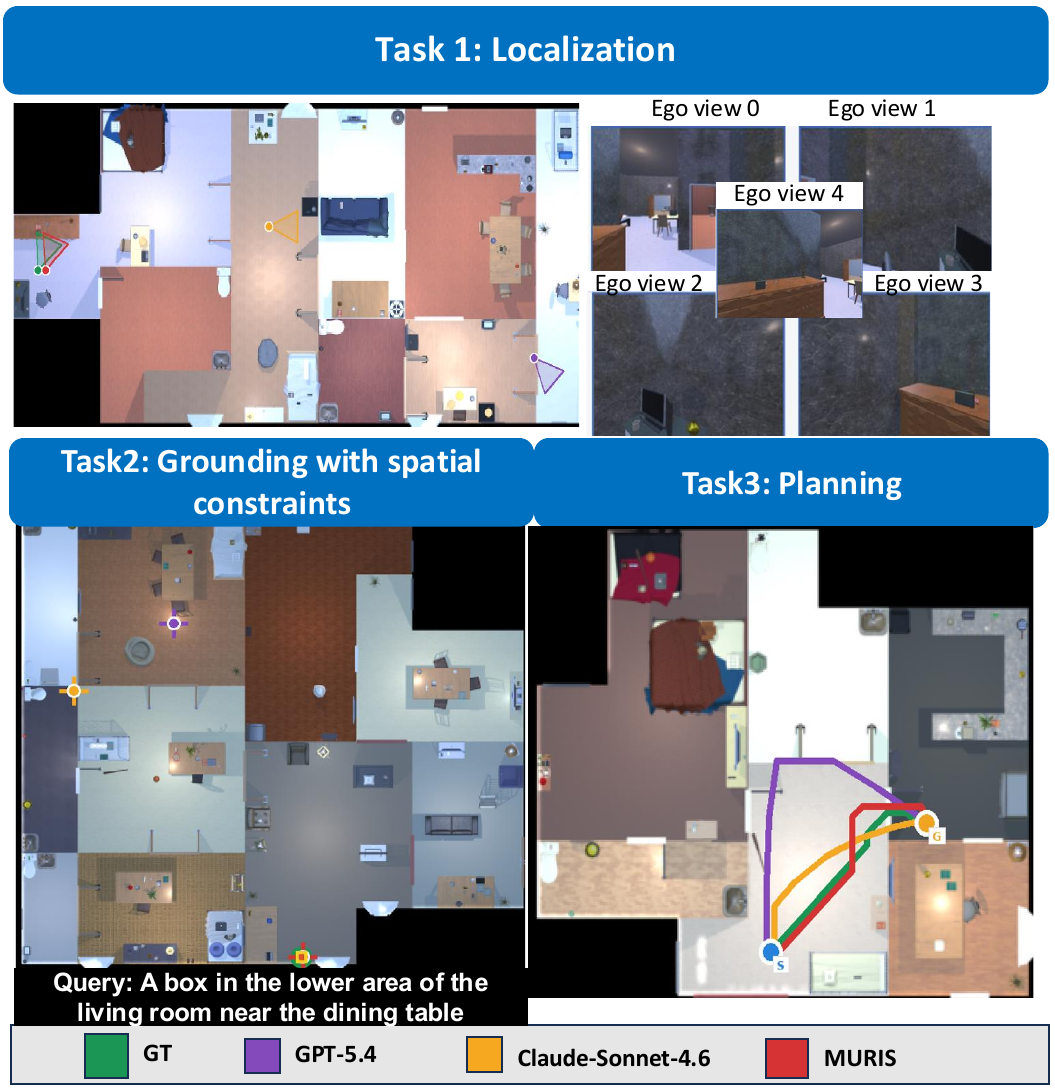}
\captionof{figure}{\textbf{Synthetic comparison with GPT-5.4 and Claude-Sonnet-4.6.} The Task~2 query is ``a box in the lower area of the living room near the dining table.'' The baselines fail to preserve the compositional spatial constraints and show large Task~1 pose errors. Their Task~3 paths make unnecessary detours, whereas \muris better matches the reference geometry.}
\label{fig:qualitative-syn-gpt-claude}

\vspace{1mm}

\includegraphics[width=0.74\linewidth]{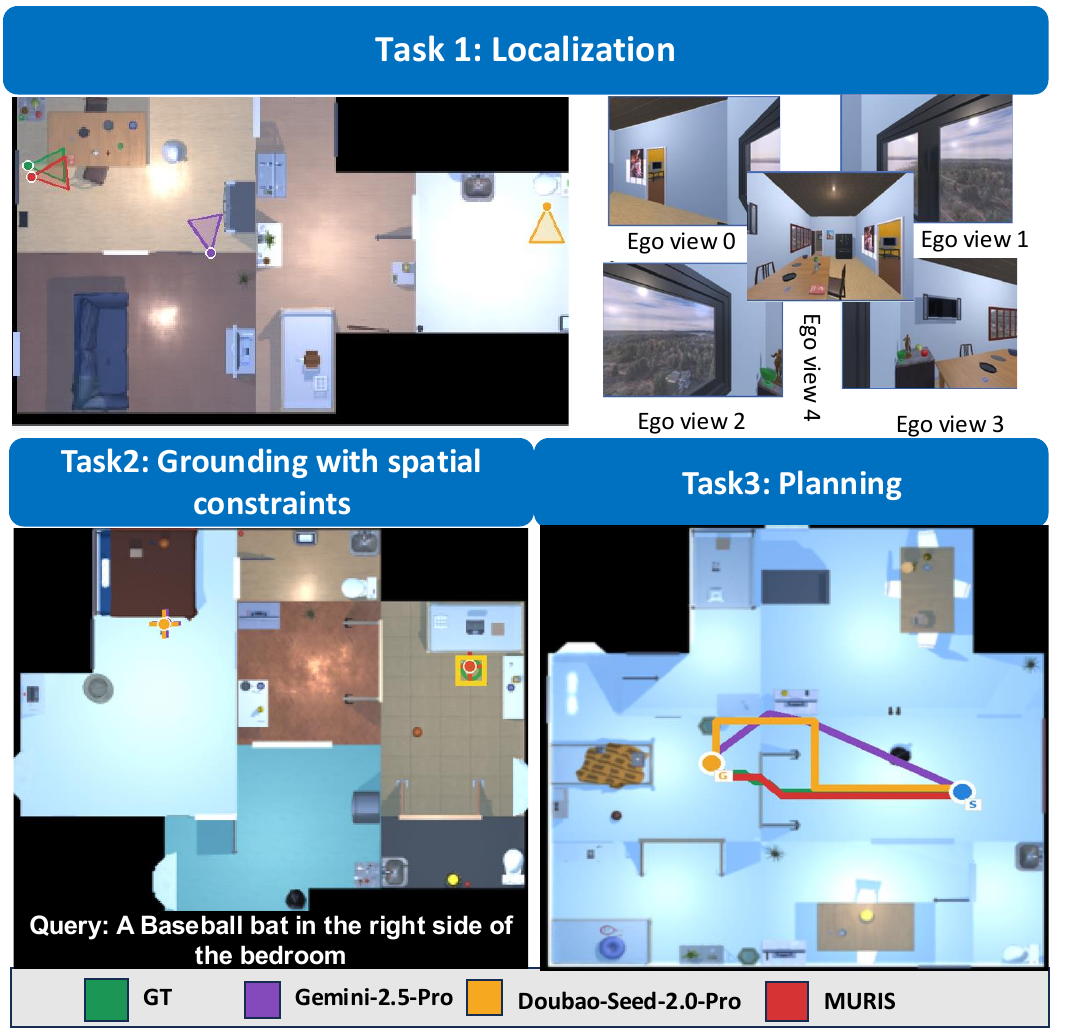}
\captionof{figure}{\textbf{Synthetic comparison with Gemini-2.5-Pro and Doubao-Seed-2.0-Pro.} The Task~2 query is ``a baseball bat on the right side of the bedroom.'' Both baselines confuse the target with other bedroom regions and substantially mislocalize the Task~1 pose. Their Task~3 routes depart from the reference path, while \muris remains aligned with it.}
\label{fig:qualitative-syn-gemini-doubao}
\end{minipage}
\end{minipage}
}]

\clearpage
\section{Additional Experimental Details}

\subsection{Reproducibility Details}
\label{app:reproducibility}

\paragraph{Controlled model variants.}\hspace{-0.6em}
The primary \muris model is trained on the union of the frozen synthetic and scan-derived splits. We evaluate this model and both single-source controls on each benchmark domain with the shared training configuration in Table~\ref{tab:training-config}.

\begin{table}[H]
\centering
\scriptsize
\setlength{\tabcolsep}{2.5pt}
\renewcommand{\arraystretch}{0.96}
\begin{tabularx}{\columnwidth}{p{0.24\columnwidth}X}
\toprule
\textbf{Component} & \textbf{Final reference-model configuration} \\
\midrule
Backbone and adaptation & Qwen3-VL-4B-Instruct with LoRA on all linear projections; rank 256, $\alpha=16$, dropout 0.05; no task-specific architectural modules. \\
Training data & Examples from \instructsyn, \instructscan, or their union. \\
Optimization & 2 epochs; fused AdamW; peak learning rate $2\times10^{-4}$; cosine schedule; 0.03 warmup ratio; 0.01 weight decay; gradient-norm clipping at 0.3. \\
Batching & 8 GPUs, 2 examples per GPU, and 2 gradient accumulation steps for an effective batch size of 32. \\
Randomization & Training examples shuffled with a fixed random seed of 42. \\
Software & PyTorch 2.8.0 with CUDA 12.8; Transformers 4.57.6; TRL 0.24.0; PEFT 0.17.1; Accelerate 1.6.0. \\
Compute & Eight NVIDIA RTX A6000 48-GB GPUs. \\
\bottomrule
\end{tabularx}
\caption{Shared training configuration for the single-source and mixed-source \muris model variants.}
\label{tab:training-config}
\end{table}

\paragraph{Inference protocol.}
Task~1 receives the RGB map followed by five ego-view images from the same camera location; Tasks~2 and~3 receive one RGB map. We do not crop the map around the queried location or target, and each base checkpoint's processor configuration is loaded without overriding its image-size parameters. No test-time voting, self-consistency sampling, or task-specific visual preprocessing is used.

\paragraph{Model-specific output conventions.}
The common benchmark interface uses normalized $[0,1000]^2$ coordinates. Molmo-7B-D and Molmo-72B-0924 are prompted in their native $[0,100]^2$ percentage-coordinate convention; every predicted spatial coordinate is then multiplied by 10 before applying the common evaluator. Orientation values are not rescaled. MolmoPoint-8B retains the common task wording and adds its native point-token trigger for Task~2. These deterministic adaptations change only the output representation, not the images, sample set, or scoring thresholds.

\subsection{Evaluation Details}
\label{app:prompts}

Figure~\ref{fig:evaluation-prompts} shows the zero-shot user-message templates and specifies the order of the visual inputs. Except for the model-specific coordinate and output adaptations above, all zero-shot baselines receive task-equivalent wording and the same output constraints.

\begin{center}
\centering
\begin{promptpanel}[listing options={style=promptbody,basicstyle=\rmfamily\scriptsize},fonttitle=\bfseries\scriptsize,left=1mm,right=1mm,top=0.8mm,bottom=0.7mm,before skip=0mm,after skip=1.2mm]{Task 1: Ego-to-Map Localization}
You are given a top-down RGB map of an indoor environment followed by five ego-view images (ego0 through ego4) captured from the same camera location.

Estimate the shared camera location and the heading of the final ego view (ego4).
Return one line only, with no explanation.
- (x, y) is the camera location in full-map normalized [0, 1000] coordinates. x increases to the right and y increases downward.
- Orientation is the heading of ego4 in degrees in [0, 360), measured clockwise from the top of the map. The headings 0, 90, 180, and 270 point up, right, down, and left, respectively.

Format: Location: (x, y), Orientation: theta
\end{promptpanel}
\begin{promptpanel}[listing options={style=promptbody,basicstyle=\rmfamily\scriptsize},fonttitle=\bfseries\scriptsize,left=1mm,right=1mm,top=0.8mm,bottom=0.7mm,before skip=0mm,after skip=1.2mm]{Task 2: Language Target Grounding}
Please act as a professional indoor object grounding assistant. You are provided with a top-down RGB map of an indoor environment.

**Task:**
Locate the object referred to by the natural language query on this map.

**Output Rules:**
- Return exactly one coordinate pair.
- The coordinate must be the center of the target object.
- Do not include any explanation.

**Coordinate System:**
- Coordinates are defined on this full map.
- Coordinates are normalized to the range [0, 1000].

**Query:** {query}

Please answer in the format: Coordinates: (x, y)
\end{promptpanel}
\begin{promptpanel}[listing options={style=promptbody,basicstyle=\rmfamily\scriptsize},fonttitle=\bfseries\scriptsize,left=1mm,right=1mm,top=0.8mm,bottom=0.7mm,before skip=0mm,after skip=0mm]{Task 3: Map-Conditioned Path Planning}
You are a careful path planning assistant for a top-down RGB indoor map.

Use the visible floor layout to infer where the traversable indoor space is. Avoid walls, furniture, obstacles, and forbidden regions. The straight line connecting any two consecutive waypoints must stay inside traversable space.

Coordinate system: Coordinates are normalized to the range [0, 1000].

Start Point: {start_point}
Goal Point: {goal_point}

Generate a waypoint path from Start to Goal. The first waypoint must be exactly the start and the last waypoint must be exactly the goal. Insert enough intermediate waypoints so each straight segment is collision-free.

Return only the final path, with no explanation:
Path: [[x1, y1], [x2, y2], ..., [xn, yn]]
\end{promptpanel}
\captionof{figure}{\textbf{Zero-shot evaluation prompt templates.} Task 1 receives a top-down RGB map followed by five ego-view images (\texttt{ego0}--\texttt{ego4}) from the same camera location. Tasks 2 and 3 each receive one top-down RGB map.}
\label{fig:evaluation-prompts}
\end{center}

\paragraph{Parsing and aggregation.}
Table~\ref{tab:parse-success} reports parse success for the output formats in Figure~\ref{fig:evaluation-prompts}. Deterministic format normalization accepts common separators and \texttt{<point>} tags without changing predicted values. Scoring uses coordinates without boundary clipping and wraps headings modulo $360^\circ$. For the success rates in Section~\ref{sec:tasks-metrics} and relaxed Path@$k$, unparseable outputs count as failures. Mean coordinate errors and median DTW use successfully parsed predictions. We pool 200 easy, 200 medium, and 100 hard synthetic examples per task, and 30 easy and 120 medium scan-derived examples per task.

\paragraph{Task 3 Endpoint and Collision Evaluation.}
\label{app:task3-evaluation}

We rasterize predicted polylines at approximately one-pixel intervals in the native map. Obstacle pixels in the binary traversability mask and out-of-bounds samples count as collisions. ProcTHOR masks use the simulator's default agent footprint and 0.25~m navigation grid. HM3D reference paths maintain at least 0.10~m obstacle clearance during generation. Scoring uses the original masks. The relaxed Path@$k$ diagnostic applies the endpoint check from Section~\ref{sec:tasks-metrics} and allows up to $k$ segments with an obstacle or out-of-bounds sample.

\paragraph{Annotation alignment.}
A shared per-scene transform aligns pose labels, target annotations, and reference waypoints. We check start points, goal points, and reference waypoints against the paired traversability mask during dataset construction.

\section{Data Details}

\subsection{Dataset Distribution}
\label{app:distribution}

\begin{figure*}[p]
\centering
\includegraphics[width=0.96\textwidth]{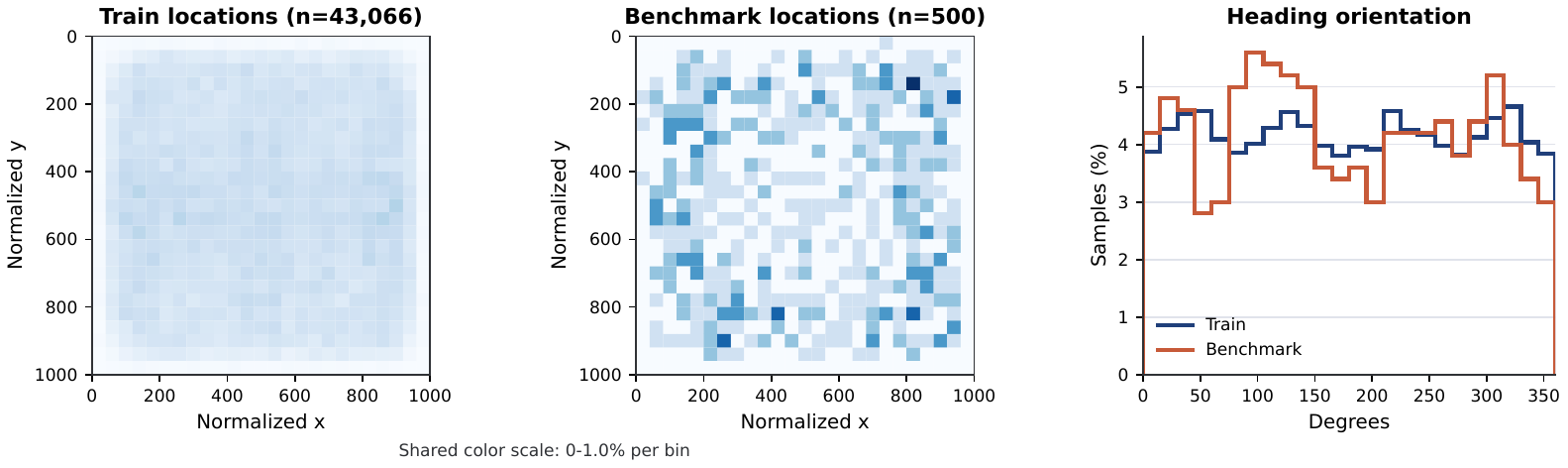}
\caption[Final synthetic Task~1 distribution]{Task~1 pose distribution for \instructsyn (43,066 samples) and \benchsyn (500). The first two panels report within-split percentages in a shared-scale \(25\!\times\!25\) grid over normalized map coordinates; the final panel uses 15\textdegree{} heading bins. Both splits cover the full map canvas, with nearly identical coordinate medians and similar mean headings (178.9\textdegree{} versus 175.8\textdegree{}).}
\label{fig:proc-distribution-task1}

\vspace{4pt}

\includegraphics[width=0.96\textwidth]{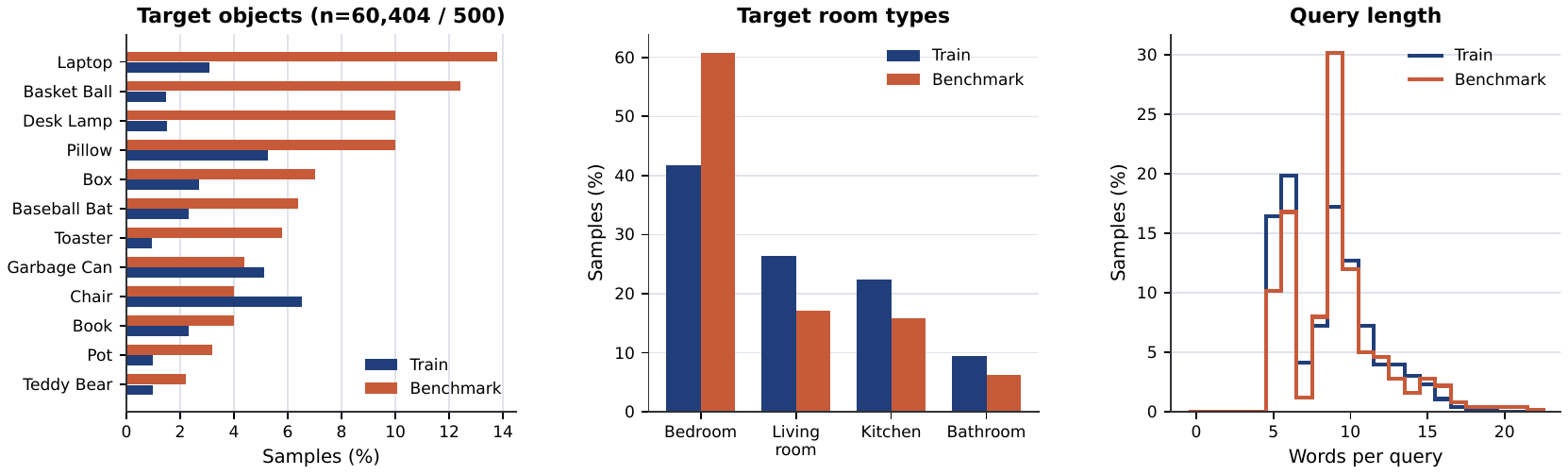}
\caption[Final synthetic Task~2 distribution]{Task~2 semantic and language distributions for \instructsyn (60,404 queries) and \benchsyn (500). The left panel compares the 12 most frequent benchmark target categories, the middle panel compares all four room types, and the right panel reports query length; each split is normalized independently. The benchmark intentionally emphasizes fine-grained targets and bedrooms, while mean query length remains similar to training (9.11 versus 8.57 words).}
\label{fig:proc-distribution-task2}

\vspace{4pt}

\includegraphics[width=0.90\textwidth]{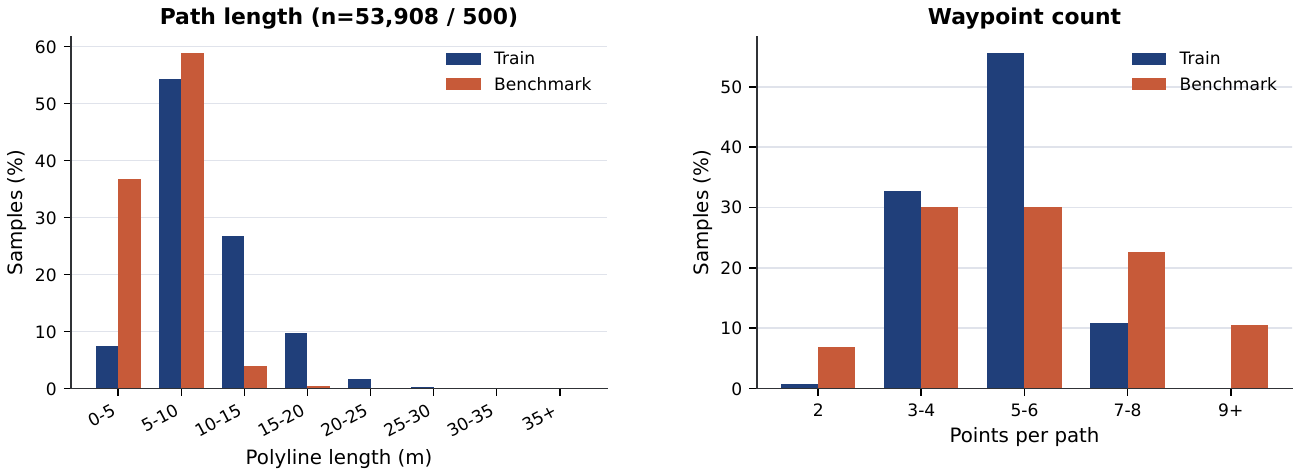}
\caption[Final synthetic Task~3 distribution]{Task~3 trajectory distributions for \instructsyn (53,908 paths) and \benchsyn (500). The panels compare within-split percentages over 5 m physical path-length bins and waypoint-count buckets. Training paths are longer on average (8.47 m versus 5.41 m), whereas the benchmark contains more 0--5 m paths and a larger share of trajectories with seven or more waypoints.}
\label{fig:proc-distribution-task3}
\end{figure*}

\begin{figure*}[p]
\centering
\includegraphics[width=0.96\textwidth]{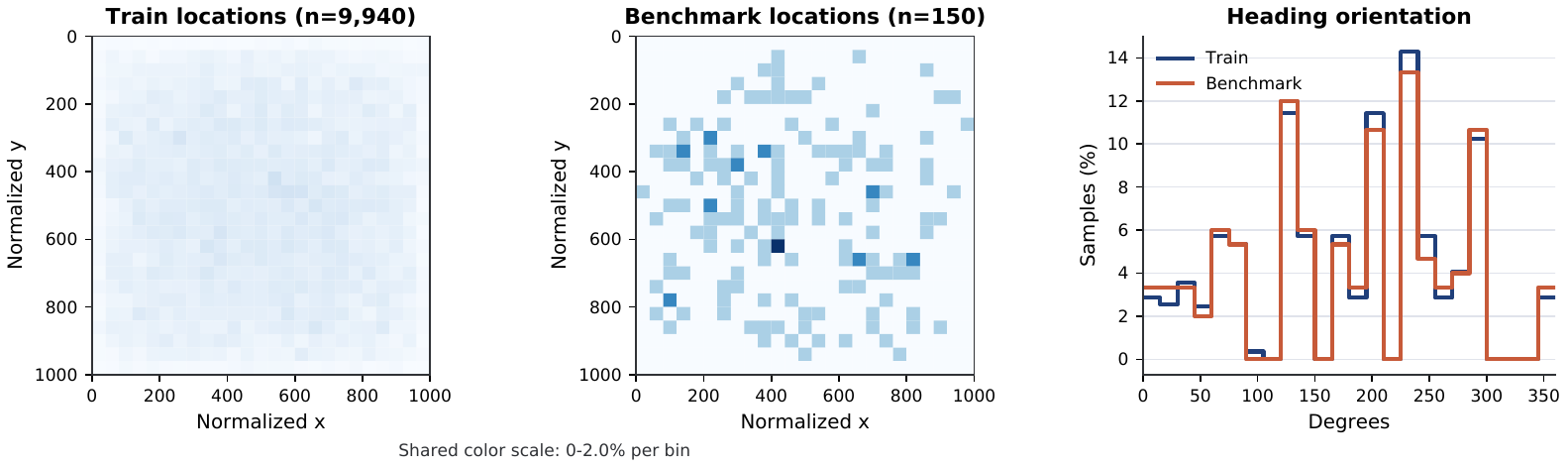}
\caption[Scan-derived Task~1 distribution]{Task~1 pose distribution for \instructscan (9,940 samples) and \benchscan (150). The location panels use within-split percentages in a shared-scale \(25\!\times\!25\) grid, and the heading panel uses 15\textdegree{} bins. Both splits span the normalized map canvas and have closely aligned heading statistics (mean 179.5\textdegree{} versus 178.2\textdegree{}), despite the smaller benchmark split.}
\label{fig:hm3d-distribution-task1}

\vspace{4pt}

\includegraphics[width=0.96\textwidth]{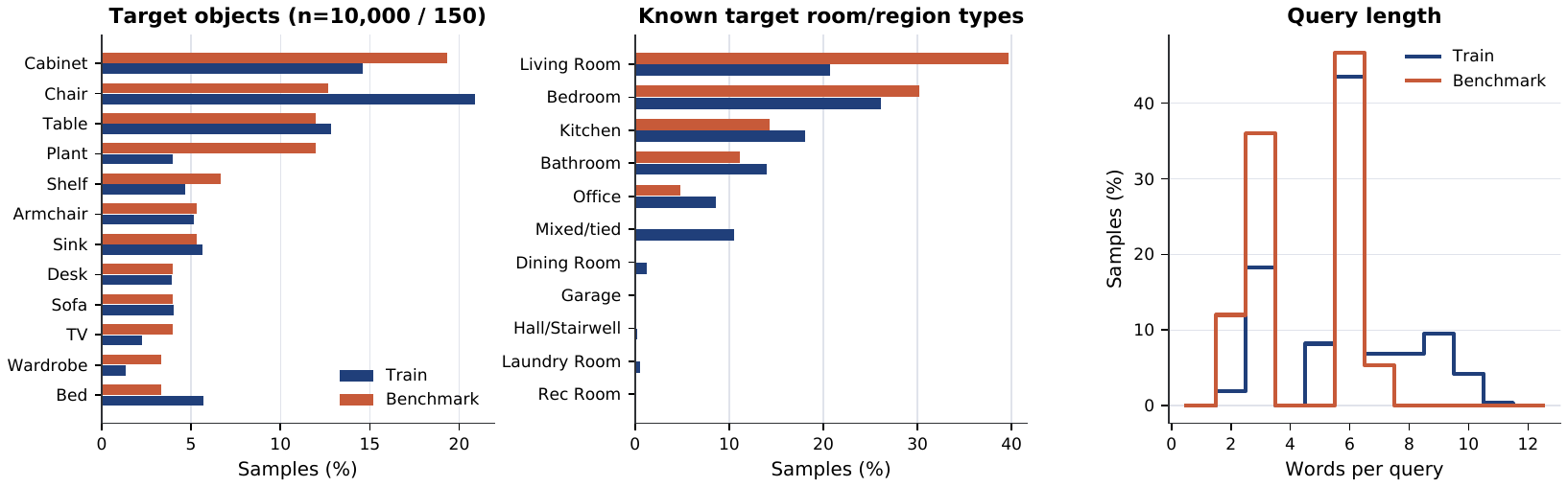}
\caption[Scan-derived Task~2 distribution]{Task~2 semantic and language distributions for \instructscan (10,000 queries) and \benchscan (150). Panels compare the 12 most frequent benchmark targets, known room or region groups, and query length using within-split percentages. The middle panel excludes missing or unknown region annotations and renormalizes each split over known labels. Relative to training, the benchmark places more weight on cabinets and plants and uses shorter queries on average (4.49 versus 5.98 words).}
\label{fig:hm3d-distribution-task2}

\vspace{4pt}

\includegraphics[width=0.90\textwidth]{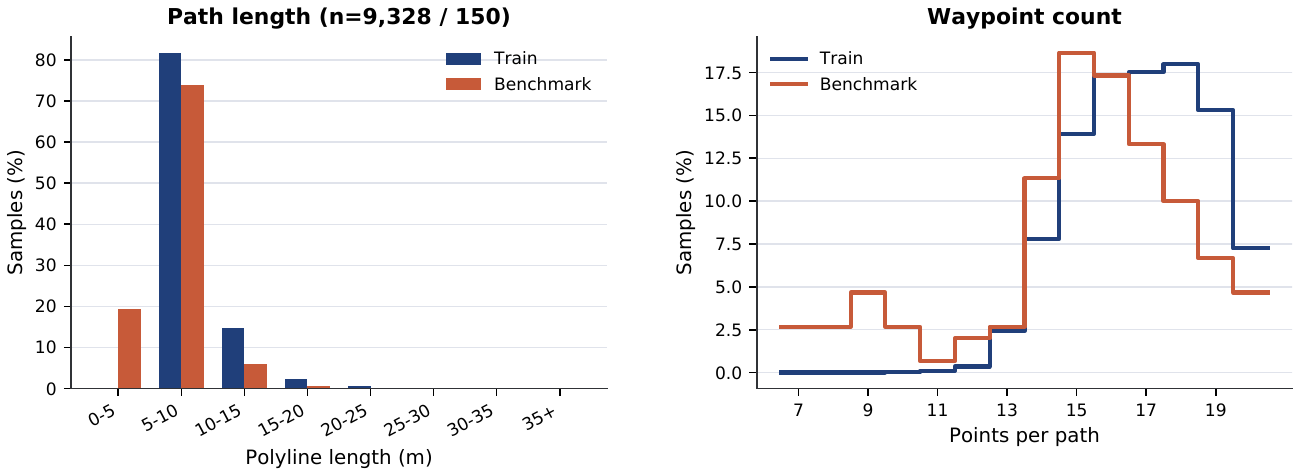}
\caption[Scan-derived Task~3 distribution]{Task~3 trajectory distributions for \instructscan (9,328 paths) and \benchscan (150). The panels report within-split percentages over 5 m physical path-length bins and exact waypoint counts. Training trajectories are longer (8.06 m versus 6.24 m on average) and contain slightly more waypoints (16.90 versus 15.17), while the benchmark includes an additional concentration of short 0--5 m paths.}
\label{fig:hm3d-distribution-task3}
\end{figure*}

Figures~\ref{fig:proc-distribution-task1}--\ref{fig:proc-distribution-task3} and~\ref{fig:hm3d-distribution-task1}--\ref{fig:hm3d-distribution-task3} compare training and evaluation distributions separately by source. Frequencies are normalized within each split unless specified otherwise.

The synthetic training split spans 37,849 rooms. In the scan-derived training split, the 59 scenes with HM3DSem labels contain 713 annotated regions. The other scan training scenes contribute localization and planning examples. In the synthetic benchmark, mean path lengths across the easy, medium, and hard tiers are 2.60, 6.34, and 9.15~m, with mean waypoint counts of 3.76, 5.84, and 8.45. The easy and medium scan-derived tiers average 3.29 and 6.98~m.

\makeatletter
\setlength{\@dblfptop}{0pt}
\setlength{\@dblfpsep}{8pt plus 2pt minus 2pt}
\setlength{\@dblfpbot}{0pt plus 1fil}
\makeatother

\onecolumn
\subsection{Data Visualization}
\label{app:visualization}

\noindent
\begin{minipage}{\textwidth}
\centering
\includegraphics[width=0.64\linewidth,trim=16bp 16bp 16bp 13bp,clip]{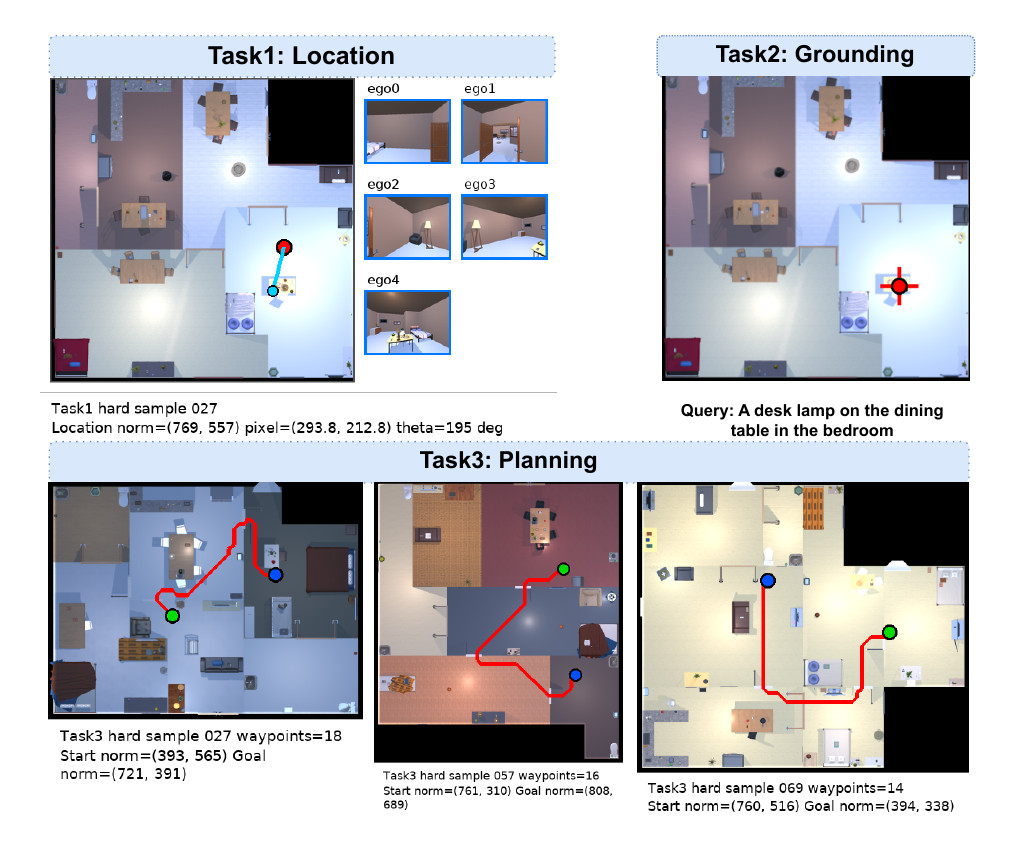}
\captionof{figure}[ProcTHOR synthetic-data visualization]{Representative ProcTHOR examples and supervision for the three tasks. Task~1 pairs a top-down RGB map with five ego-view observations sampled at one pose; the pose overlay encodes the target location and heading. Task~2 associates a referring expression with one target-instance center. Task~3 shows waypoint polylines connecting the supplied start and goal across different route complexities. Coordinates are normalized to \([0,1000]^2\). Overlays in both figures show ground truth.}
\label{fig:proc-task-visualization}

\vspace{2mm}

\includegraphics[width=0.64\linewidth,trim=16bp 13bp 17bp 10bp,clip]{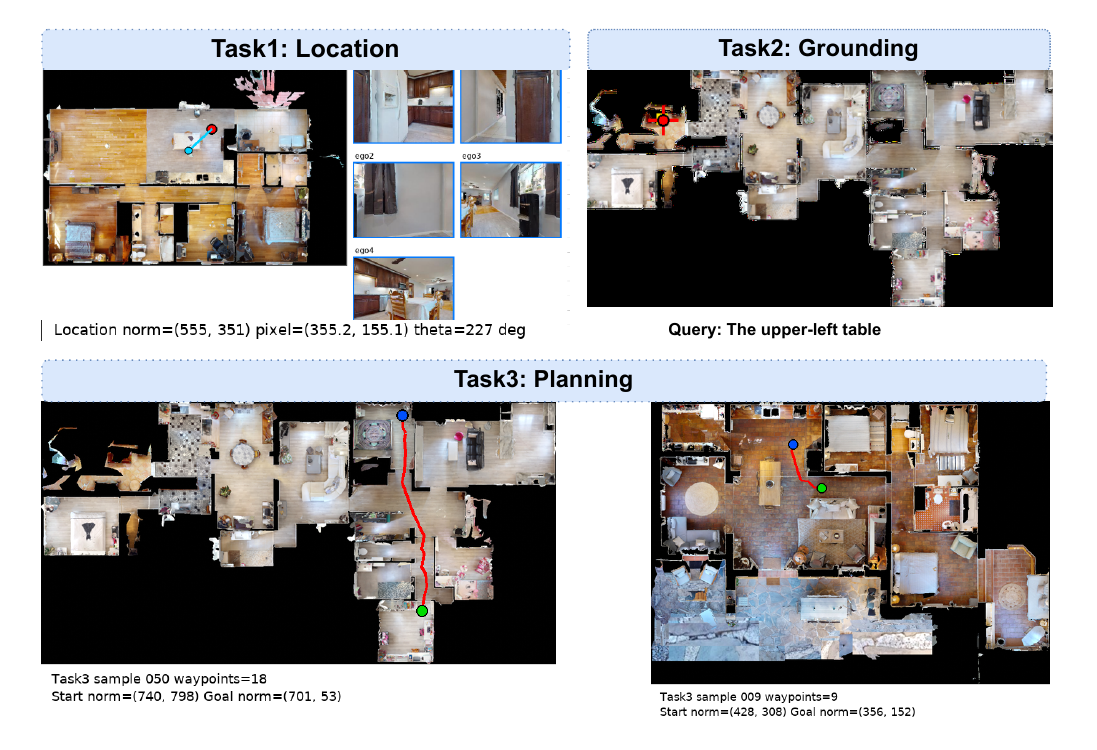}
\captionof{figure}[HM3D scan-data visualization]{Representative HM3D examples and supervision for the three tasks. Reconstructed orthographic RGB maps retain irregular boundaries, partial observations, and scan artifacts. Task~1 pairs one map with five ego views, Task~2 grounds a spatial referring expression to an instance center, and Task~3 represents a route as an ordered waypoint sequence from start to goal. Both data sources expose the same normalized \([0,1000]^2\) output interface, while metric errors are computed after mapping predictions back to each native map raster.}
\label{fig:hm3d-task-visualization}
\end{minipage}